\documentclass[sigconf]{acmart}
\usepackage{enumitem}
\usepackage[noend]{algpseudocode}
\usepackage{algorithmicx,algorithm}
\usepackage{float}
\usepackage{balance}

\AtBeginDocument{%
  \providecommand\BibTeX{{%
    \normalfont B\kern-0.5em{\scshape i\kern-0.25em b}\kern-0.8em\TeX}}}

\setcopyright{acmlicensed}
\copyrightyear{2018}
\acmYear{2018}
\acmDOI{XXXXXXX.XXXXXXX}

\acmConference[MM'24]{Make sure to enter the correct
  conference title from your rights confirmation email}{October 28 - November 1,
  2024}{Melbourne, Australia.}
\acmISBN{978-1-4503-XXXX-X/18/06}

\begin{document}
\title{PlacidDreamer: Advancing Harmony in Text-to-3D Generation}


\author{Shuo Huang}
\affiliation{%
  \institution{Tsinghua University}
  \city{Beijing}
  \country{China}}
\email{huangs22@mails.tsinghua.edu.cn}

\author{Shikun Sun}
\affiliation{%
  \institution{Tsinghua University}
  \city{Beijing}
  \country{China}}
\email{ssk21@mails.tsinghua.edu.cn}

\author{Zixuan Wang}
\affiliation{%
  \institution{Tsinghua University}
  \city{Beijing}
  \country{China}}
\email{wangzixu21@mails.tsinghua.edu.cn}

\author{Xiaoyu Qin}
\affiliation{%
  \institution{Tsinghua University}
  \city{Beijing}
  \country{China}}
\email{xyqin@tsinghua.edu.cn}

\author{Yanmin Xiong}
\affiliation{%
  \institution{Kuaishou Technology}
  \city{Beijing}
  \country{China}}
\email{xiongyanmin@kuaishou.com}

\author{Yuan Zhang}
\affiliation{%
  \institution{Kuaishou Technology}
  \city{Beijing}
  \country{China}}
\email{zhangyuan03@kuaishou.com}

\author{Pengfei Wan}
\affiliation{%
  \institution{Kuaishou Technology}
  \city{Beijing}
  \country{China}}
\email{wanpengfei@kuaishou.com}

\author{Di Zhang}
\affiliation{%
  \institution{Kuaishou Technology}
  \city{Beijing}
  \country{China}}
\email{zhangdi08@kuaishou.com}

\author{Jia Jia}
\authornote{Corresponding author.}
\affiliation{%
  \institution{Tsinghua University}
  \institution{Beijing National Research Center for Information Science and Technology}
  \city{Beijing}
  \country{China}}
\email{jjia@tsinghua.edu.cn}
\renewcommand{\shortauthors}{Shuo Huang et al.}

\begin{abstract}
Recently, text-to-3D generation has attracted significant attention, resulting in notable performance enhancements. 
Previous methods utilize end-to-end 3D generation models to initialize 3D Gaussians, multi-view diffusion models to enforce multi-view consistency, and text-to-image diffusion models to refine details with score distillation algorithms.
However, these methods exhibit two limitations. 
Firstly, they encounter conflicts in generation directions since different models aim to produce diverse 3D assets. 
Secondly, the issue of over-saturation in score distillation has not been thoroughly investigated and solved.
To address these limitations, we propose PlacidDreamer, a text-to-3D framework that harmonizes initialization, multi-view generation, and text-conditioned generation with a single multi-view diffusion model, while simultaneously employing a novel score distillation algorithm to achieve balanced saturation.
To unify the generation direction, we introduce the Latent-Plane module, a training-friendly plug-in extension that enables multi-view diffusion models to provide fast geometry reconstruction for initialization and enhanced multi-view images to personalize the text-to-image diffusion model.
To address the over-saturation problem, we propose to view score distillation as a multi-objective optimization problem and introduce the Balanced Score Distillation algorithm, which offers a Pareto Optimal solution that achieves both rich details and balanced saturation.
Extensive experiments validate the outstanding capabilities of our PlacidDreamer. 
The code is available at \url{https://github.com/HansenHuang0823/PlacidDreamer}.
\end{abstract}





\maketitle


\begin{figure*}[ht]
\begin{center}
\centerline{\includegraphics[width=0.95\textwidth]{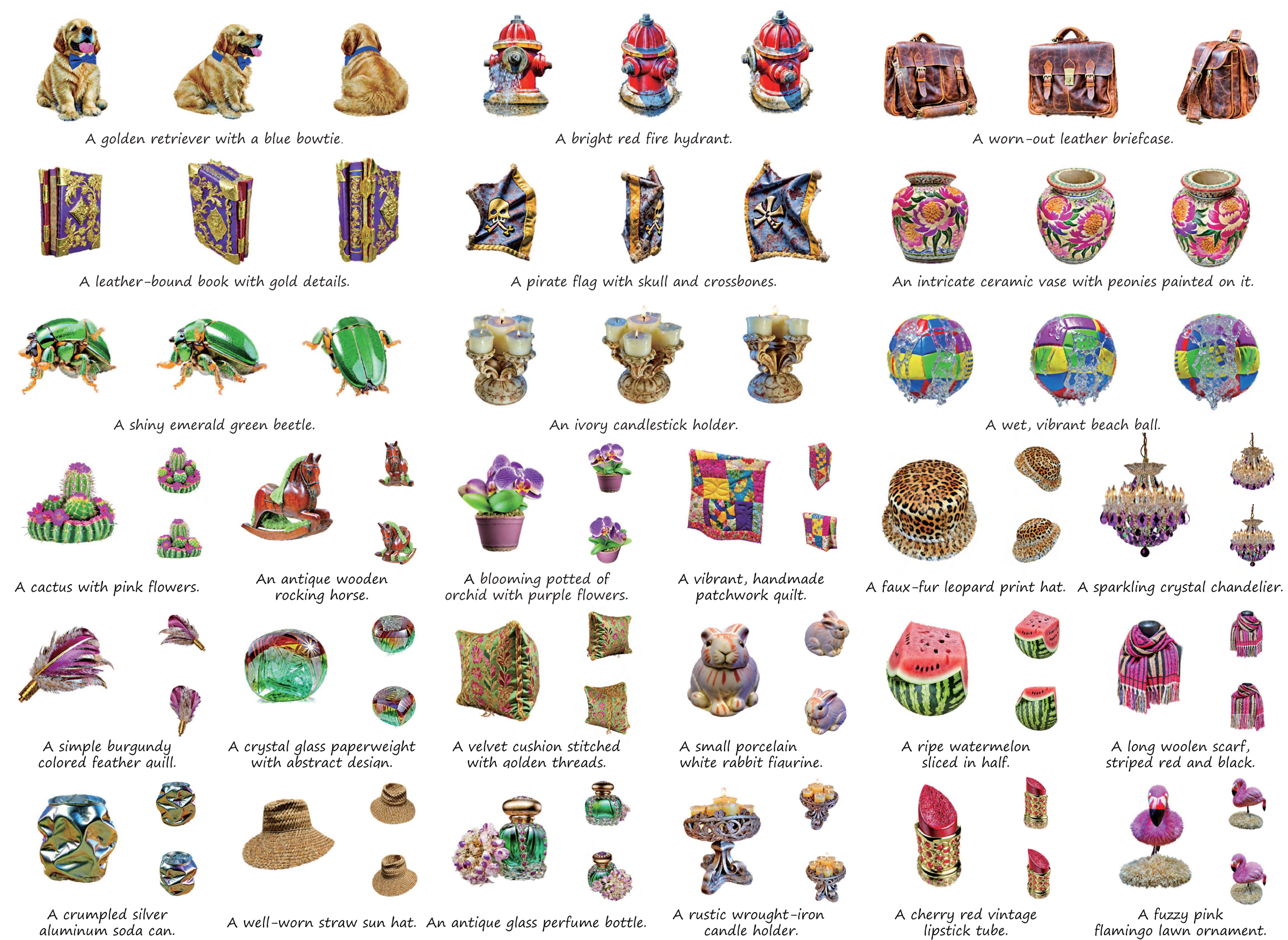}}
\vskip -0.1in
\caption{3D generations of PlacidDreamer.}
\label{fig:introduction}
\end{center}
\vskip -0.3in
\end{figure*}

\section{Introduction}
The task of generating 3D assets from text, known as text-to-3D, has garnered significant attention for its potential to simplify 3D creation, a process once requiring specialized knowledge.
Due to the relative scarcity and lower quality of 3D data compared with 2D data, one promising approach is to adapt pre-trained 2D models for 3D generation. An optimization-based approach leveraging score distillation algorithms, which distill generative capability from pre-trained 2D diffusion models to guide subsequent 3D generations, has emerged as a dominant approach in this field.

Since the introduction of the first score distillation algorithm, Score Distillation Sampling (SDS) \cite{poole2022dreamfusion}, subsequent works have significantly advanced this optimization-based approach, enhancing both generation quality and speed. One significant factor influencing generation quality is the multi-face problem. To address this problem, Magic123 \cite{qian2023magic123}, DreamCraft3D \cite{sun2023dreamcraft3d}, Consistent123 \cite{lin2023consistent123}, and EfficientDreamer \cite{zhao2023efficientdreamer} incorporate multi-view diffusion models to enhance multi-view consistency. More recently, the introduction of 3D Gaussian Splatting \cite{kerbl20233d} has further optimized the pipeline with convenient initialization, faster rendering, and training speed. LucidDreamer \cite{EnVision2023luciddreamer}, GaussianDreamer \cite{yi2023gaussiandreamer}, and GSGEN \cite{chen2023text} propose to leverage end-to-end 3D generation models \cite{jun2023shap, nichol2022point} to provide a robust 3D Gaussian initialization, thereby enhancing overall quality.

\begin{figure*}[t]
\begin{center}
\centerline{\includegraphics[width=\textwidth]{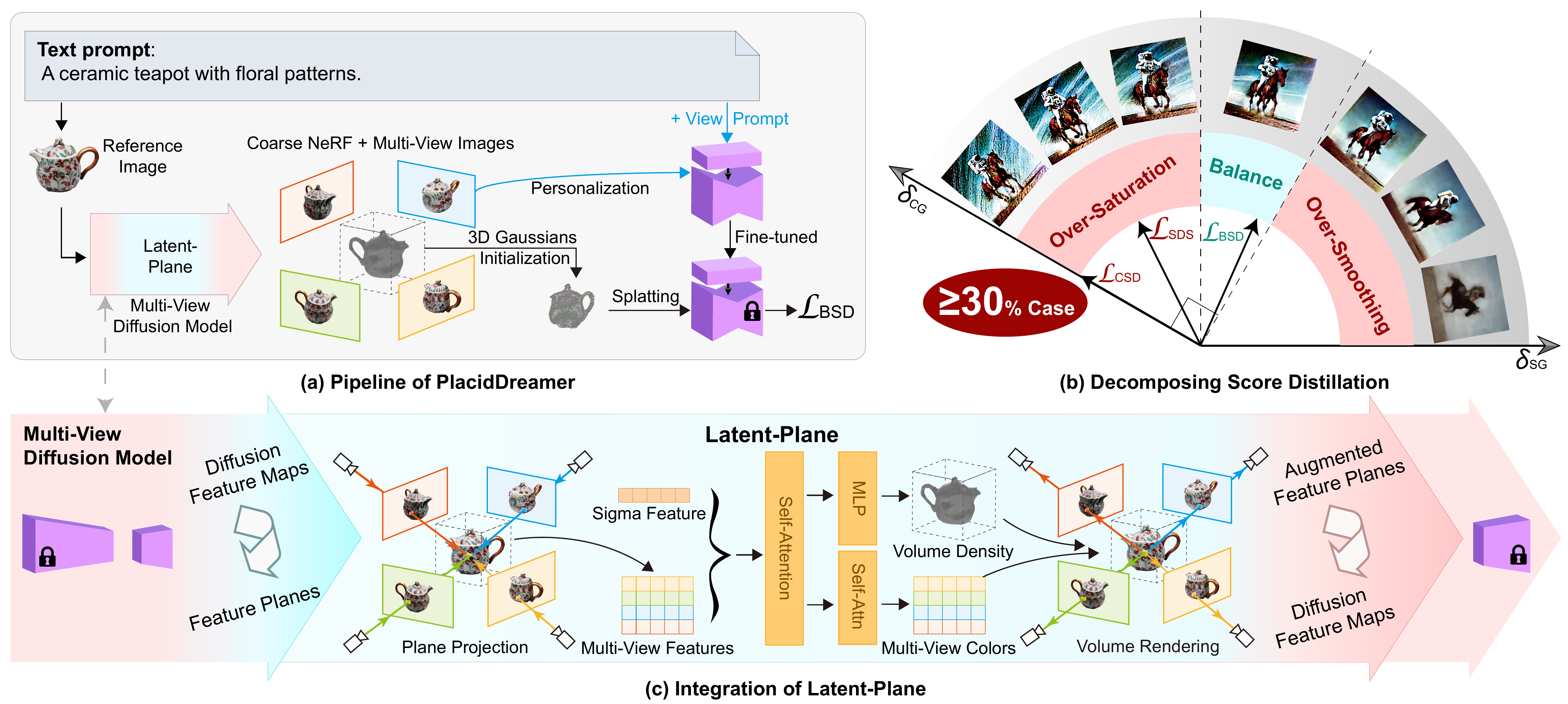}}
\vskip -0.1in
\caption{(a) The pipeline of PlacidDreamer. (b) Score distillation can be decomposed into two directions: classifier guidance $\delta_\mathrm{CG}$ and smoothing guidance $\delta_\mathrm{SG}$. CSD \cite{yu2023text} only utilizes classifier guidance. In more than 30\% of cases, the angle between these two guidance vectors is obtuse. In such scenarios, using a fixed CFG parameter in SDS may result in negative optimization in the $\delta_\mathrm{SG}$ direction, leading to over-saturation. However, BSD algorithm ensures that each optimization step is non-negative in both directions. (c) The integration of the Latent-Plane module with multi-view diffusion models.}
\label{fig:methods}
\end{center}
\vskip -0.3in
\end{figure*}

Despite significant progress in the aforementioned methods, there are still two main limitations that require attention: 

\begin{itemize}[leftmargin=*]
\item[--]{\bfseries Conflicting Optimization Directions.} The integration of multiple generative models within a single pipeline can lead to contradictory optimization directions. 
For instance, the score distillation guidance derived from multi-view diffusion models may be at odds with that from text-to-image diffusion models \cite{qian2023magic123}, necessitating the development of specialized balancing strategies \cite{lin2023consistent123} or additional refinement \cite{sun2023dreamcraft3d} for optimal performance.
Moreover, employing distinct generative models for supervision across different stages \cite{yi2023gaussiandreamer, chen2023text} can cause the generative model in later stages to ignore the outputs of its predecessors and independently generate new results based on its intrinsic data interpretations.

\item[--]{\bfseries Over-Saturation in Score Distillation.} The problem of over-saturation within score distillation algorithms remains insufficiently explored and unresolved.
This issue manifests as a discrepancy in color distribution between the 3D content created through score distillation techniques and the 2D images generated by diffusion inference processes.
Although certain methodologies yield results with more appropriate levels of saturation, they impose a significant computational load. 
Examples include LoRA finetuning \cite{wang2023prolificdreamer}, adversarial training \cite{chen2023it3d}, or supervision in multiple spaces with iterative sampling \cite{zhu2023hifa}. Thus, a comprehensive understanding of the over-saturation problem and a fast yet effective score distillation algorithm is needed.
\end{itemize}

To overcome these challenges, we propose PlacidDreamer, a framework that harmonizes initialization, multi-view generation, and text-conditioned generation with a single multi-view diffusion model, while simultaneously employing a novel score distillation algorithm to achieve balanced saturation.
More specifically, to address the first limitation, we newly devised a Latent-Plane module. This module enhances the multi-view diffusion model by enabling fast geometry reconstruction and improving its capabilities in generating multi-view images. The reconstructed geometry is then utilized to initialize 3D Gaussian points, and the improved multi-view images are used to personalize the text-to-image diffusion model with directional prompts. This coordinated approach aligns the generation directions within the pipeline with the outcomes from the multi-view diffusion model, promoting convergence and significantly enhancing the quality of the generated content.
Additionally, the Latent-Plane module is designed to be training-efficient and is adaptable to a variety of multi-view diffusion models, accommodating different viewpoint configurations. 
To address the second limitation, we delve into the causes of over-saturation. We decompose the score distillation equation into two primary components: classifier guidance and smoothing guidance. Prior algorithms have suffered from an imbalance where classifier guidance overwhelmingly dominates smoothing guidance, leading to prevalent over-saturation.
Our analysis reveals that in over 30\% of instances, the optimization directions of these two guidances form obtuse angles, resulting in negative optimization along certain directions with a fixed Classifier-Free Guidance (CFG) parameter, making it challenging to maintain control over balance after multiple optimization steps.
To rectify this imbalance, we propose to treat score distillation as a multi-objective optimization problem and introduce a Balanced Score Distillation (BSD) algorithm. This algorithm incorporates a multi-objective optimization solution, Multiple-Gradient Descent Algorithm (MGDA) \cite{desideri2012multiple}, which dynamically adjusts the optimization directions to converge at Pareto Optimal points, where the generated results exhibit both rich details and balanced saturation.

We evaluate PlacidDreamer both qualitatively and quantitatively to demonstrate its effectiveness. Extensive experiments indicate the superiority of our method over previous methods. Quantitative evaluations conducted on the T3Bench \cite{he2023t} benchmark reveal that PlacidDreamer consistently outperforms baseline methods by a margin of at least five points in both generation quality and alignment metrics. Furthermore, we conduct ablation studies to respectively evaluate the contribution of each proposed module to the overall quality. To further highlight the BSD algorithm's effectiveness, we replace the score distillation algorithms in various open-source text-to-3D frameworks with BSD, while keeping other components constant. Results show that BSD consistently enhances these frameworks' performance.

Our contributions can be summarized as follows:
\begin{itemize}
    \item 
    We introduce PlacidDreamer, a novel framework for high-fidelity text-to-3D generation. PlacidDreamer advances a harmonious generation process through two novel approaches: the Latent-Plane module, enhancing multi-view diffusion models, and the Balanced Score Distillation algorithm, enabling richer details and saturation control.
    
    \item
    We identify the causes of the over-saturation problem in score distillation algorithms and propose to view score distillation as a multi-objective optimization problem, with the goal of optimizing towards Pareto Optimal points to stabilize the outcomes of generation.

    \item 
    Extensive experiments, including both quantitative metrics and qualitative assessments, demonstrate that PlacidDreamer significantly outperforms existing state-of-the-art methods.
    
\end{itemize}

\section{Related Works}
\textbf{Text-to-3D Generation.}
DreamFields \cite{jain2022zero} initially employs a pre-trained model, CLIP \cite{radford2021learning}, to guide the optimization of NeRF \cite{mildenhall2021nerf}. Trying to leverage the generative nature of the diffusion model, Dreamfusion \cite{poole2022dreamfusion} introduces SDS, a loss associated with a score function derived from distilling 2D diffusion models. Subsequent works significantly improve SDS-based methods, including those using multi-view consistent models \cite{shi2023mvdream, li2023sweetdreamer, zhao2023efficientdreamer, liu2023syncdreamer}, enhancing pipeline structures \cite{lin2023magic3d, chen2023fantasia3d, yi2023gaussiandreamer, wu2024hd, metzer2023latent, sargent2023zeronvs, lorraine2023att3d}, introducing extra generation priors \cite{huang2023avatarfusion, tang2023stable, xu2023seeavatar, huang2023dreamwaltz, jiang2023avatarcraft, zeng2023avatarbooth, cao2023dreamavatar}, or exploring timestep scheduling \cite{huang2023dreamtime}. Recently, some fast-forward reconstruction models \cite{hong2023lrm,liu2023one,long2023wonder3d,tang2023dreamgaussian,liu2023syncdreamer} have emerged for faster 3D generation.

\noindent\textbf{Score Distillation.}
Given the heavy reliance on score distillation methods in these studies, addressing associated issues with SDS is crucial. While many works \cite{wang2023steindreamer, tang2023stable, yu2023text, hertz2023delta, katzir2023noise, wang2023score} focus on addressing over-smoothing, fewer tackle over-saturation problems. ProlificDreamer \cite{wang2023prolificdreamer} introduces VSD, leveraging LoRA finetuning for 3D distribution modeling, which can also alleviate over-saturation. IT3D \cite{chen2023it3d} handles over-saturation by training a discriminator distinguishing 3D assets from text-to-2D images. Additionally, HiFA \cite{zhu2023hifa} presents an iterative score distillation process for a more accurate sampling direction. It's worth noting that these methods introduce a significant computational burden, slowing down overall speed.

\section{Preliminaries}
\subsection{Diffusion Models}
For discrete-time diffusion models~\cite{sohl-dickstein15,ho2020denoising,nichol2021improved, dhariwal2021diffusion}, given a data distribution $q(\mathbf{x}) = q_0(\mathbf{x}_0)$, we construct a forward process with a series of distributions $q_t(\mathbf{x}_t) = \mathcal{N}(\alpha_t \mathbf{x}_0, (1-\alpha_t^2)\mathbf{I})$ with decreasing $\{\alpha_t | t \in [0,T], t \in \mathbb{Z}\}$, where $\alpha_0 = 1$ and $\alpha_T \approx 0$. In the generation process, we start from $\mathbf{x}_T \sim \mathcal{N}(\mathbf{0}, \mathbf{I})$ and generate a sample of the previous timestep iteratively with a denoising network $\mathbf{\epsilon}_\phi(\mathbf{x}_t, t)$ trained by minimizing the prediction of added noise, which is given by
\begin{equation}
    \label{equ:diffusion loss}
    \mathbb{E}_{\mathbf{x}_0 \sim q_0(\mathbf{x}_0), t, \mathbf{\epsilon} \sim \mathcal{N}(\mathbf{0}, \mathbf{I})} w(t)\|\mathbf{\epsilon}_\phi(\alpha_t\mathbf{x}_0 + \sigma_t \mathbf{\epsilon}, t) - \mathbf{\epsilon}\|_2^2,
\end{equation}
where $w(t)$ is to balance losses between different timesteps, $t$ is uniformly selected from $0$ to $T$, and $\sigma_t = \sqrt{1 - \alpha_t^2}$.
\subsection{Score Distillation Sampling (SDS)}
Score distillation is an optimization-based generation method that distills knowledge from pre-trained 2D diffusion models to guide other generations, The first score distillation method, SDS \cite{poole2022dreamfusion} is denoted as
\begin{equation}
\label{equ:SDS}
    \nabla_{\theta}{\mathcal{L}_{\mathrm{SDS}}}{(\phi, \mathbf{x}=g(\theta))} \triangleq \mathbb{E}_{t,\epsilon}[\omega(t)(\hat{\epsilon}_{\phi}(\mathbf{x}_{t};y,t)-\epsilon) \frac{\partial\mathbf{x}}{\partial\theta}],
\end{equation}
where $\mathbf{x}_t = \alpha_t \mathbf{x} + \sigma_t \mathbf{\epsilon}$ and other symbols are defined the same as in Equation~\eqref{equ:diffusion loss}.
Intuitively, this loss perturbs $\mathbf{x}$ with a random amount of noise corresponding to the timestep t, and estimates an update direction that follows the score function of the diffusion model to move to a higher density region.


\subsection{Multiple-Gradient Descent Algorithm} 
MGDA \citet{desideri2012multiple} is a gradient-based algorithm used to solve multi-objective optimization problems. Multi-objective optimization refers to the task of finding a Pareto Optimal solution under multiple optimization criteria, where optimizing one objective does not deteriorate the solution of another objective during the optimization process.
In the context of solving multi-objective optimization tasks, MGDA determines a descent direction that is common to all criteria. For instance, when applied to a binary-objective optimization problem optimizing $\mathcal{L}^1(x)$ and $\mathcal{L}^2(x)$, MGDA will identify the direction orthogonal to $\nabla_x\mathcal{L}^1(x) - \nabla_x\mathcal{L}^2(x)$. The algorithm has been demonstrated to converge to a Pareto Optimal point, where there is no available optimization that can improve one objective without worsening another.

\section{Methods}
\label{sec:methods}
\subsection{Pipeline}
Given a text prompt, we initially generate a reference image using Stable Diffusion \cite{rombach2022high} or MVDream \cite{shi2023mvdream}. Unlike previous methods that transform text-to-3D generation into single-view reconstruction, we do not aim to precisely reconstruct the reference object. After background removal, the reference image is fed into the multi-view diffusion model to generate multi-view images. To optimize the initialization of 3D Gaussian points \cite{kerbl20233d} and ensure its compatibility with the multi-view images, we introduce Latent-Plane. This module, as detailed in Section~\ref{subsec:latent_plane}, seamlessly integrates into the latent layers of any multi-view diffusion model, enabling fast reconstruction of a volume density field and enhancement of the generated images within 40 seconds. Moreover, it is training-friendly (only requiring 4 $\times$ A6000 GPUs for 18 hours), as it directly leverages 3D spatial knowledge from pre-trained multi-view diffusion model. Subsequently, the volume density field is used to initialize the 3D Gaussian points on the object surface, while the generated multi-view images are used to fine-tune Stable Diffusion with LoRA \cite{hu2021lora} technique. During this fine-tuning process, directional prompts such as "left view" are incorporated into the original text, enhancing the diffusion model's awareness of 3D space. Finally, we supervise the splatted images using the BSD guidance from the fine-tuned diffusion model. As elaborated in Section~\ref{subsec:BSD}, we decompose score distillation and reveal a conflict between optimization directs, leading to color discrepancies. We suggest treating score distillation as a multi-objective optimization problem and we introduce the BSD algorithm, achieving rich details and reasonable colors.

\subsection{Latent Plane}
\label{subsec:latent_plane}

The proposed method, Latent-Plane, serves as a plug-in module compatible with various multi-view diffusion models.
These diffusion models typically take an input image $I_{\pi_0}$ of an object from a specific viewpoint $\pi_0$ and generate corresponding images $I_{\pi_i}$ from other viewpoints $\pi_i, i=1,2,\ldots, N$.
Inspired by ConsisNet \cite{yang2023consistnet}, which reinforces feature patches by utilizing coordinate relationships derived from back-projection, we hypothesize that pre-trained features within multi-view diffusion models possess sufficient knowledge of the 3D space to conduct direct reconstruction.
Leveraging this idea, we treat latent feature maps akin to the feature plane in the Tri-Plane \cite{chan2022efficient} model, allowing us to reconstruct a volume density field. Furthermore, the reconstruction results can be utilized to enhance the latent feature map through volume rendering, thereby improving the accuracy of multi-view predictions.

\subsubsection{Multi-View Feature Gathering}
Most multi-view diffusion models currently utilize the Unet \cite{ronneberger2015u} architecture for their denoisers. To illustrate our strategy of selecting the appropriate latent layer for inserting the Latent-Plane module, we consider the Unet architecture as an exemplar. Within the Unet architecture, assuming there are $L$ decoder blocks generating $L$ feature maps denoted as $\mathbf{F}_j \in \mathbb{R}^{H \times W \times D}$ with $j$ indexing these maps and $D$ representing the feature dimension. We choose the feature map with the highest resolution and the smallest index, denoted as $\mathbf{F}_k$, as it encompasses the deepest and most intricate features extracted by the model, containing comprehensive geometric information.

At each timestep $t$, during the reverse diffusion process, we obtain $N$ feature maps $\mathbf{F} ^ {(i)}_k, i = 1,2,\ldots, N$ from $N$ viewpoints. For every point $\mathbf{x}$ in the 3D space, we project it onto each feature map $\mathbf{F}_k^{(i)}$ and derive its corresponding feature $\mathbf{f}_k^{(i)}(\mathbf{x})$ through tri-linear interpolation, denote as,
\begin{equation}
    \mathbf{f}^{(i)}(\mathbf{x}) = \mathrm{Interp\_2D}(\mathrm{Proj}(\mathbf{x}, \mathbf{F}_k^{(i)}), \mathbf{F}_k^{(i)}),
\end{equation}
where $\mathrm{Interp\_2D}$ represents the 2D interpolation function, and $\mathrm{Proj}$ denotes the function for projecting 3D spatial points onto 2D planes. We then apply an additional linear layer to extract low-dimensional features from the extracted features. To better represent the features of spatial point $\mathbf{x}$, we concatenate each feature with its respective camera embeddings $\mathbf{e}^{(i)}_\mathrm{cam}$ and coordinate embeddings $\mathbf{e}_\mathrm{pos}(\mathbf{x})$. 
\begin{equation}
    \mathbf{e}^{(i)}(\mathbf{x}) = \mathrm{Concat}[\mathrm{Linear}(\mathbf{f}^{(i)}(\mathbf{x})), \mathbf{e}^{(i)}_\mathrm{cam}, \mathbf{e}_\mathrm{pos}(\mathbf{x})].
\end{equation}

\subsubsection{Plane-Based NeRF}
So far, $\mathbf{x}$ has $N$ features gathered from $N$ viewpoints, serving as $N$ tokens, which will be mutually enhanced through the Attention layers. To obtain the volume density $\sigma(\mathbf{x})$ for the spatial point, before feeding into the attention layer, we additionally add a token $\tau_{\sigma}$ corresponding to obtaining sigma. $\tau_{\sigma}$ is obtained by concatenating a trainable embedding $\mathbf{e}_{\sigma}$ with embeddings of timestep $t$, followed by a linear layer.
\begin{equation}
    \tau_\sigma = \mathrm{Linear}(\mathrm{Concat}[\mathbf{e}_\sigma, \mathbf{e}_t]),
\end{equation}

\begin{equation}
    \tau(\mathbf{x}) = \mathrm{Concat}[\tau_\sigma, \mathbf{e}^{(1)}(\mathbf{x}), \mathbf{e}^{(2)}(\mathbf{x}), \ldots, \mathbf{e}^{(N)}(\mathbf{x})],
\end{equation}

\begin{equation}
    \tau'(\mathbf{x}) = \mathrm{MultiHeadSelfAttn}(\tau(\mathbf{x})),
\end{equation}
where $\mathrm{MultiHeadSelfAttn}$ denotes the Multi-Head Self Attention layers.
After the multi-view augmentation, there are a total of N+1 tokens in $\tau'(\mathbf{x})$. For the sigma token $\tau'_\sigma(\mathbf{x})$, we use an MLP to obtain its volume density value $\sigma(\mathbf{x})$.
For the multi-view enhanced feature tokens $\tau'(\mathbf{x})$, we conduct volume rendering to gather features enhanced on the object surface.
\begin{equation}
\label{equ:sigma}
    \sigma(\mathbf{x}) = \mathrm{MLP}(\tau'_\sigma(\mathbf{x})).
\end{equation}
\begin{equation}
    {\mathbf{F}}'^{(i)}_k=\mathrm{Volume\_Rendering}(\sigma(\mathbf{x}), \tau'^{(i)}(\mathbf{x}))
\end{equation}
We directly add the feature map to the original feature map. The enhanced feature map continues to function in the decoder of the diffusion model, ultimately producing the model's predicted noise $\epsilon$, which is then integrated into diffusion training and inference.
\begin{figure*}[t]
\begin{center}
\centerline{\includegraphics[width=0.95\textwidth]{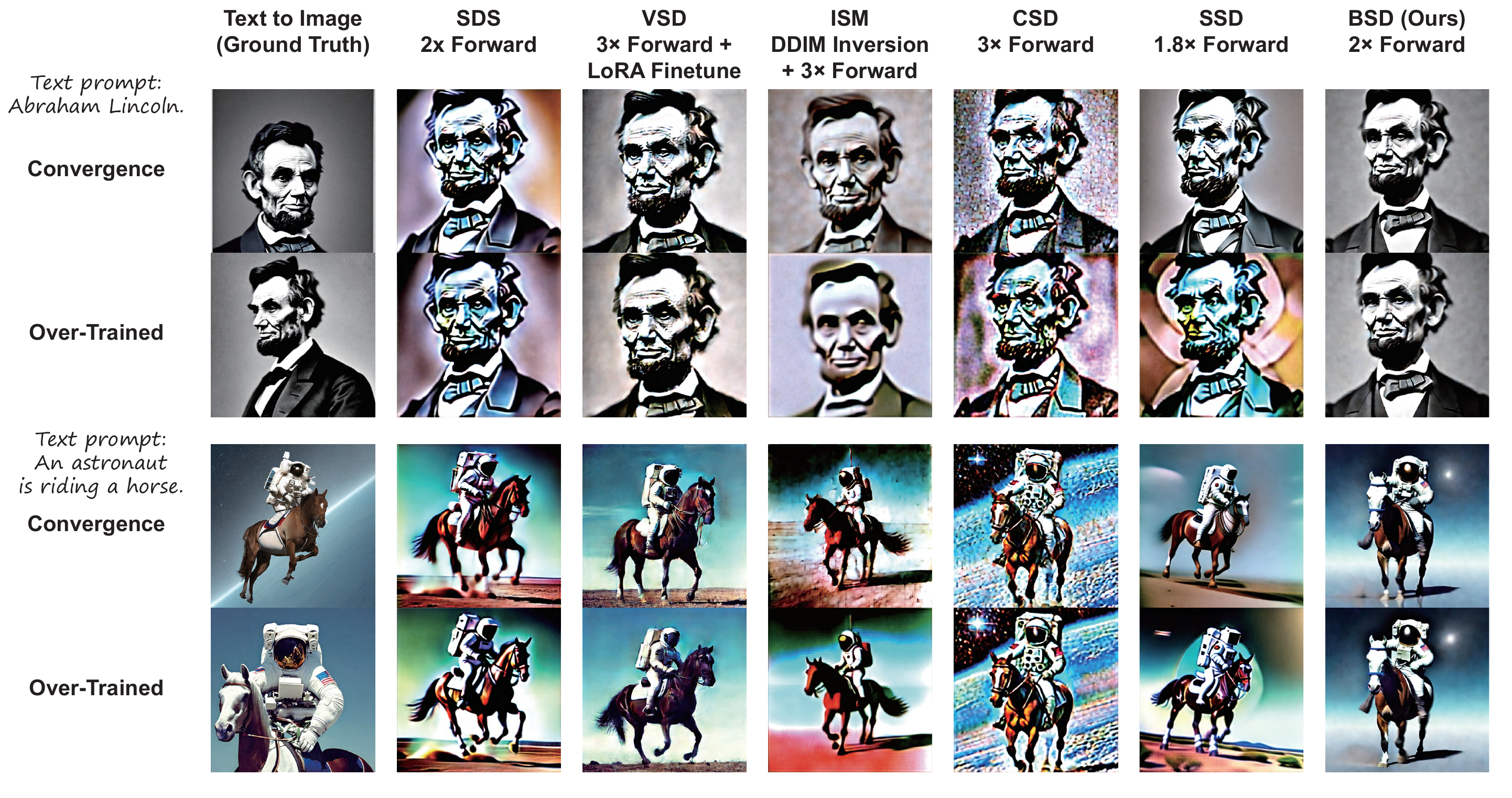}}
\vskip -0.2in
\caption{2D generation results of score distillation algorithms, annotated with computational costs. "Forward" represents undergoing one forward process of the diffusion model. The results of our BSD closely resemble the text-to-image ground truth. BSD converges at the Pareto Optimal points, ensuring that its results maintain balanced saturation during over-training.}
\label{fig:BSD_2D_exp}
\end{center}
\vskip -0.3in
\end{figure*}
\subsubsection{Training.}
The training comprises two stages. In the first stage, we independently train the volume density generation module, enabling the selection of arbitrary camera viewpoints, rather than being constrained to predefined fixed camera viewpoints. Subsequently, we can calculate the depth of each spatial point relative to the selected camera. Following volume rendering, we supervise the generated occupancy map $\mathrm{M}^{(i)}$ and depth maps $\mathrm{D}^{(i)}$, which are denoted as
\begin{equation}
    \mathcal{L}_\sigma = \mathcal{L}_\mathrm{BCE}(\mathrm{M}^{(i)}, \mathrm{M}^{(i)}_\mathrm{gt}) + \lambda (1-\rho(\mathrm{D}^{(i)}, \mathrm{D}^{(i)}_\mathrm{gt})),
\end{equation}
where, $\mathcal{L}_\mathrm{BCE}$ denotes the Binary Cross Entropy loss, $\rho$ represents the Pearson correlation coefficient, and $\lambda$ controls the balance between the two losses.
For training the feature map augmentation, we use the standard diffusion loss described in Equation~\eqref{equ:diffusion loss}.

\subsection{Balanced Score Distillation}
\label{subsec:BSD}
In this chapter, our exploration of score distillation is conducted through 2D generation experiments, where we directly optimize the pixel values of a 2D image. The use of 2D experiments allows the exclusion of various factors unrelated to score distillation algorithms, such as camera viewpoints and 3D representation selection, thereby highlighting the effectiveness of score distillation methods. Furthermore, the outcomes of 2D experiments can effectively represent the effects observed in 3D experiments~\cite{wang2023prolificdreamer, hertz2023delta}.
Our focus on the experimental results is specifically directed toward the richness of details and color saturation at two distinct time points: the optimal state of image quality (referred to as the convergence state) and the period of sustained training after reaching the convergence state (referred to as the over-trained state). In 3D generation, the lack of an early-stop mechanism frequently results in the widespread occurrence of over-training.

\subsubsection{Decomposing Score Distillation}
We adopt a modeling approach different from SDS \cite{poole2022dreamfusion} to obtain our decomposition. Score distillation is an optimization-based generation method aimed at distilling knowledge from pre-trained 2D diffusion models to guide other generations.
In order to utilize 2D diffusion models for guidance, it is necessary to establish a connection between the 2D image distribution $p_0$ modeled by diffusion models and the 3D representation distribution $q$.
We make the assumption that the probability density distribution $q(\theta | y)$ for the parameters $\theta$ of the 3D representation, conditioned on the text prompt $y$, is proportional to the product of the conditional probability densities of its rendered images $\mathbf{x}_0^\pi$ under various viewpoint $\pi$s, denoted as,

\begin{equation}
q(\theta | y) \propto \prod_\pi p_0(\mathbf{x}_0^\pi(\theta)|y).
\end{equation}
We adopt the average negative logarithm of the probability,
\begin{equation}
    \mathcal{L}_\mathrm{SD} = - \frac{1}{N}\log q(\theta|y),
\end{equation}
as the loss for neural network training, where $N$ is the number of $\pi$s.
Consequently, as the loss decreases, we obtain a sample as described in the text prompt.
We can get the gradient on $\theta$ as,
\begin{equation}
    \nabla_{\theta}\mathcal{L}_\mathrm{SD} = -\mathbb{E}_\pi[\nabla_{\theta} \log p_0(\mathbf{x}_0^\pi(\theta)|y)].
\end{equation}
However, $p_0$ represents the unknown distribution of general 2D images, and high-density regions of $p_0$ are sparsely populated \cite{song2019generative}. Given that current diffusion models are designed to model the score function of the distribution $p_t$ of noisy images with noise of level $t$, we instead leverage it to generate an optimized gradient for the noisy images $\mathbf{x}_t$ that is obtained by adding noise of level $t$ to $\mathbf{x}_0^\pi$. Subsequently, this gradient can be back-propagated to the image $\mathbf{x}_0^\pi$ and further transmitted to the 3D parameters $\theta$, denoted as,
\begin{equation}
\label{equ:SD}
    \nabla_{\theta}\mathcal{L}_\mathrm{SD} = -\mathbb{E}_{\pi,t}[\nabla_{\mathbf{x}_t} \log p_t(\mathbf{x}_t(\theta)|y) \frac{\partial{\mathbf{x}_t}}{\partial{\mathbf{x}_0^\pi}} \frac{\partial{\mathbf{x}_0^\pi}}{\partial{\theta}}].
\end{equation}
Please note that we observe a distinction between our formula and Equation (\ref{equ:SDS}) from SDS \cite{poole2022dreamfusion}, particularly in the absence of the final term $-\epsilon$ in our formulation.
The first term $\nabla_{\mathbf{x}_t} \log p_0(\mathbf{x}_t(\theta)|y)$ is a 2D score function modeled by diffusion models.
In 2D diffusion generation, the score function of conditional distribution can be decomposed into the classifier guidance and the score function of an unconditional distribution:
\begin{equation}
    \nabla_{\mathbf{x}_t} \log p_t(\mathbf{x}_t|y) = \nabla_{\mathbf{x}_t} \log p_t(y|\mathbf{x}_t) + \nabla_{\mathbf{x}_t} \log p_t(\mathbf{x}_t).
\end{equation}
So, we naturally decompose the score distillation process into two functional terms.
\begin{equation}
    \delta_\mathrm{CG} = -\nabla_{\mathbf{x}_t} \log p_t(y|\mathbf{x}_t)= (\epsilon(\mathbf{x}_t, t, y) - \epsilon(\mathbf{x}_t, t, \emptyset)) / \sigma_t
\end{equation}
\begin{equation}
    \delta_\mathrm{SG} = -\nabla_{\mathbf{x}_t} \log p_t(\mathbf{x}_t)= \epsilon(\mathbf{x}_t, t, \emptyset) / \sigma_t
\end{equation}
\begin{equation}
    -\nabla_{\mathbf{x}_t} \log p_t(\mathbf{x}_t|y) = u \cdot \delta_\mathrm{CG} + v \cdot \delta_\mathrm{SG},
\end{equation}
where $u$ and $v$ control the ratio between the two terms.
The first term is the classifier guidance term ($\delta_\mathrm{CG}$). When only utilizing classifier guidance, referred to as CSD~\cite{yu2023text}, images exhibit fine details but suffer from artifacts and over-saturation, as shown in the upper right corner of Figure~\ref{fig:methods}. Therefore, the introduction of the second term aims to alleviate the artifacts caused by $\delta_\mathrm{CG}$ and guide $\mathbf{x}_0$ towards the higher density region in the distribution of 2D images. As depicted in Figure~\ref{fig:methods}, starting from $\mathcal{L}_\mathrm{CSD}$, as $u/v$ decreases, the optimization directions move towards the "balance" area, the generation results become smoother. Hence, we refer to it as smoothing guidance ($\delta_\mathrm{SG}$). However, when $\delta_\mathrm{CG}$ starts to dominate the proportion, the image tends to become overly smoothed.

\begin{figure*}[htbp]
\begin{center}
\centerline{\includegraphics[width=0.95\textwidth]{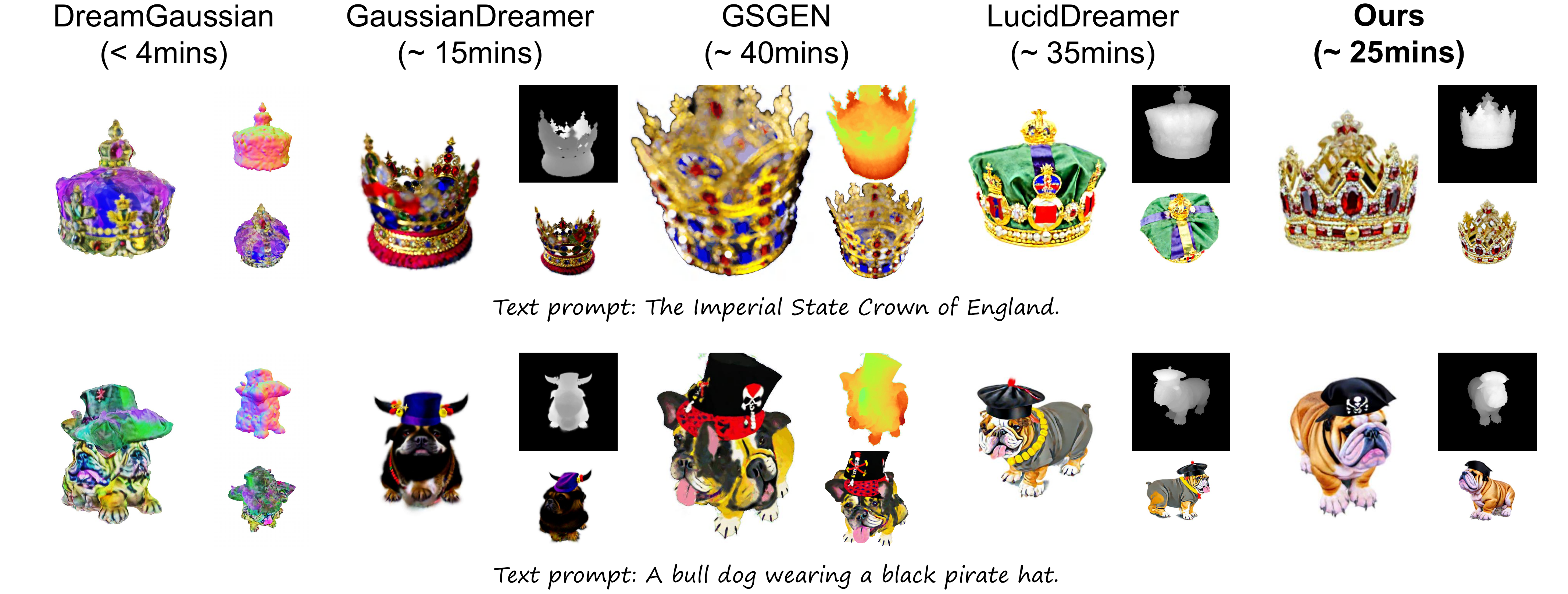}}
\vskip -0.1in
\caption{Qualitative comparison with baseline methods. More comparisons with recent methods utilizing NeRF \cite{mildenhall2021nerf} as 3D representations are provided in Appendix~\ref{sec:appendix_qualitative}.}
\label{fig:baseline}
\end{center}
\vskip -0.3in
\end{figure*}


\subsubsection{Balanced Score Distillation}
Based on our analysis, the parameters of CFG ($u/v$) in score distillation algorithms play a critical role in regulating the balance between the two guidance terms during each optimization step.
However, we have observed that their effectiveness becomes highly unstable after numerous optimization steps, particularly highlighted by the failure of most score distillation methods to preserve image properties during the over-training phase.
We conduct experiments of previous score distillation algorithms including SDS \cite{poole2022dreamfusion}, VSD \cite{wang2023prolificdreamer}, ISM \cite{EnVision2023luciddreamer}, CSD \cite{yu2023text}, SSD \cite{tang2023stable}. We find that most of them exhibit over-saturation during over-trained steps, as depicted in Figure~\ref{fig:BSD_2D_exp}.

To address this issue, we investigated the distribution characteristics of $\delta_\mathrm{CG}$ and $\delta_\mathrm{SG}$. We observed that over 30\% of the steps in the iterative optimization process exhibit a negative dot product between $\delta_\mathrm{CG}$ and $\delta_\mathrm{SG}$, indicating obtuse angles between their optimization directions in the high-dimensional space, as shown in Figure~\ref{fig:methods}. In such instances, simply combining the two directions based on a fixed ratio may lead to the final optimization direction projecting negatively onto one of the term's optimization directions. Consequently, the control of balance becomes ineffective due to the presence of negative optimization, ultimately resulting in over-saturation or over-smoothing.

Therefore, we propose to consider score distillation as a multi-objective optimization task, where the optimization objectives are $\mathcal{L}^1=-\lambda\log(p_t(y|\mathbf{x}_t))$ and $\mathcal{L}^2=-\log(p_t(\mathbf{x}_t))$, with $\lambda$ representing the hyper-parameter that controls the ratio of two guidance terms. The solution of multi-objective optimization tasks will find the Pareto optimal point, where no action can improve one objective without deteriorating another. If generation reaches the Pareto Optimal points, it will stabilize at this optimal point, and the generated content will exhibit rich details while maintaining balanced saturation. In this scenario, the hyper-parameter $\lambda$ genuinely controls the final inclination towards both optimization directions, rather than attempting to control the balance at each step.

We propose the Balanced Score Distillation, which employs MGDA \cite{desideri2012multiple}. 
Suppose the optimization combination of $\nabla_{\mathbf{x}_t}\mathcal{L}^1$ and $\nabla_{\mathbf{x}_t}\mathcal{L}^2$ is
\begin{equation}
    \nabla_{\mathbf{x}_t} \log \Tilde{p}_t(\mathbf{x}_t|c) = \alpha \nabla_{\mathbf{x}_t}\mathcal{L}^1 + (1 - \alpha) \nabla_{\mathbf{x}_t}\mathcal{L}^2.
\end{equation} 
Following~\citet{desideri2012multiple}, $\alpha$ is the solution of 
\begin{equation}
\label{equ:PofMGDA}
    \min _{\alpha, 1-\alpha \ge 0 }\left\{\left\| \alpha \nabla_{\mathbf{x_t}}\mathcal{L}^1 + (1-\alpha)\nabla_{\mathbf{x_t}}\mathcal{L}^2\right\|_2^2\right\}.
\end{equation}
For binary situation like in Equation~\eqref{equ:PofMGDA}, we have closed form solution for $\alpha$, which is $\alpha = \min (\max(0,\hat{\alpha}), 1)$, where
\begin{equation}
\label{equ:MGDA}
   \hat{\alpha} = \frac{(\nabla_{\mathbf{x}_t}\mathcal{L}^2 - \nabla_{\mathbf{x}_t}\mathcal{L}^1)^T\nabla_{\mathbf{x}_t}\mathcal{L}^2}{\|\nabla_{\mathbf{x}_t}\mathcal{L}^2 - \nabla_{\mathbf{x}_t}\mathcal{L}^1\|_2^2}.
\end{equation}

Thus, the final formula for BSD is:
\begin{equation}
    \nabla_{\theta} \mathcal{L}_\mathrm{SD} = \mathbb{E}_{\pi,t}[\omega(t)(\alpha \lambda \delta_\mathrm{CG} + (1 - \alpha) \delta_\mathrm{SG}) \frac{\partial{\mathbf{x}_0^\pi}}{\partial{\theta}}],
\end{equation}
\begin{equation}
   \alpha = \min\left[\max\left[0,\frac{(\delta_\mathrm{SG} - \lambda \delta_\mathrm{CG})^T \delta_\mathrm{SG}}{\|\delta_\mathrm{SG} - \lambda \delta_\mathrm{CG}\|_2^2}\right], 1\right].
\end{equation}

As illustrated in Figure~\ref{fig:BSD_2D_exp}, images generated by the BSD algorithm approximate the saturation of 2D images, showcasing smooth color transitions and detailed contours. In states of over-training, BSD maintains a harmonized saturation level, unlike other methods that suffer from loss of details due to over-saturation. A comparable method to BSD is VSD \cite{wang2023prolificdreamer}, which exhibits fewer instances of over-saturation and finer details compared to other baseline methods. However, as annotated in the figure, VSD requires significantly longer computational time and higher GPU utilization than BSD.

\subsubsection{Relationship with Previous Methods.}
We provide an analysis of the relationship between our decomposition and score distillation algorithm with previous works in Appendix~\ref{sec:appendix_score_distillation}.

\begin{figure*}[htbp]
\begin{center}
\centerline{\includegraphics[width=0.95\textwidth]{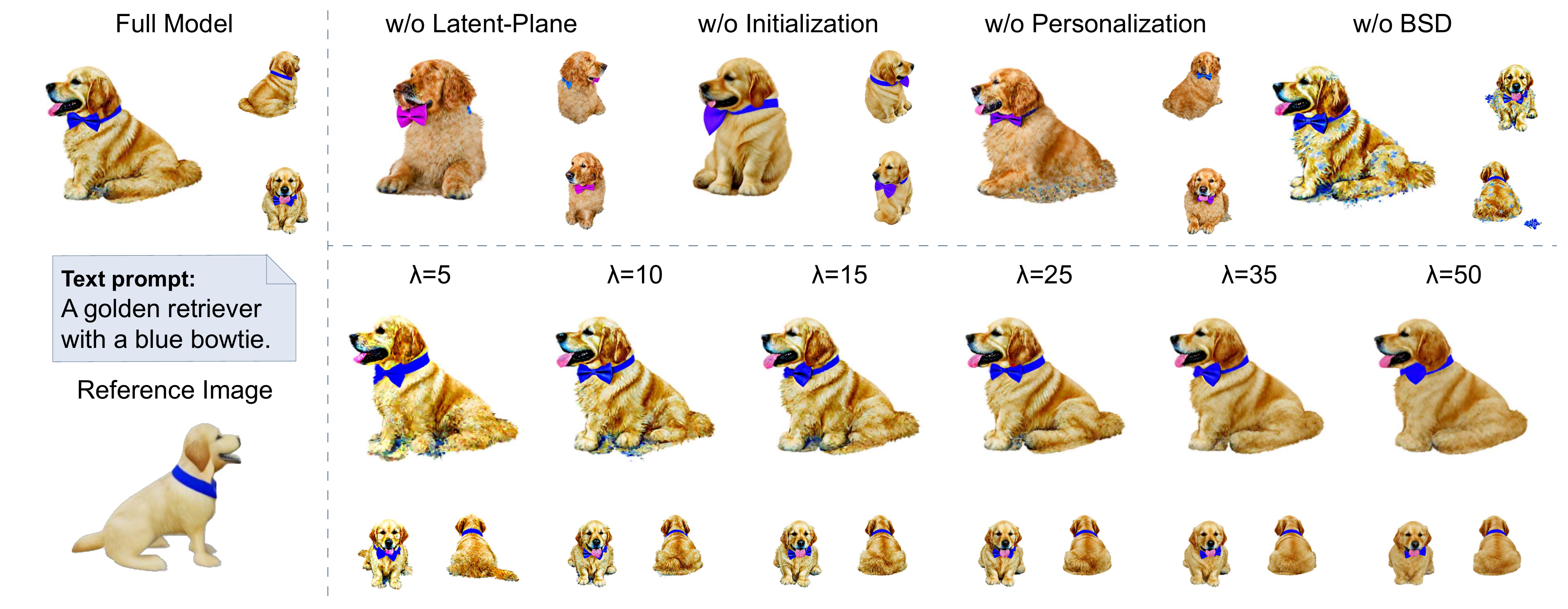}}
\vskip -0.1in
\caption{Results of ablation studies. In the first line, we evaluate PlacidDreamer by removing each component individually. In the second line, we investigate the impact of different $\lambda$ values, validating that BSD enables stable control of the balance between color saturation and detail level.}
\label{fig:ablation_study}
\end{center}
\vskip -0.3in
\end{figure*}
\section{Experiments}
\begin{table}[h]
    \centering
        \caption{Results of quantitative comparison with baselines and ablation studies. The values in parentheses represent the direct scores given by the ImageReward \cite{xu2023imagereward} model.}
    \label{tab:T3Bench}
    \vskip -0.15in
    \begin{tabular}{c|ccc}
        \hline
         & Quality & Alignment & Average \\
        \hline
        Fantasia3D (mesh) & 29.2 (-1.040) & 23.5 & 26.4 \\
        Magic3D (mesh) & 38.7 (-0.565) & 35.3 & 37.0 \\
        GeoDream (mesh) & 48.4 (-0.080) & 33.8 & 41.1 \\
        MVDream (NeRF) & 53.4 (+0.170) & 41.9  & 47.7 \\
        RichDreamer (mesh) & 57.3 (+0.365) & 40.0 & 48.6 \\
        ProlificDreamer (mesh) & 51.1 (+0.055) & 47.8 & 49.4 \\
        LucidDreamer (3DGS) & 55.7 (+0.285) & 53.8 & 54.8 \\
        \hline
        PlacidDreamer (Ours) & \textbf{62.4 (+0.620)} & \textbf{59.8} & \textbf{61.1} \\
        w/o Latent-Plane & 56.5 (+0.325) & 54.6 & 55.6 \\
        w/o Initialization & 57.3 (+0.365) & 56.1 & 56.7 \\
        w/o Personalization & 61.8 (+0.590) & 59.5 & 60.7 \\
        w/o BSD & 60.2 (+0.510) & 57.4 & 58.8 \\
        \hline
    \end{tabular}

\vskip -0.3in
\end{table}

\subsection{Implementation Details}
The Latent Plane model has $N=8$ viewpoints. For all the MLP described in Section~\ref{sec:methods}, we use a single linear layer for computing efficiency. We train the Latent-Plane module with Zero-1-to-3 \cite{liu2023zero} for 18 hours on 4 $\times$ Nvidia RTX A6000 GPUs with a batch size of 32 and loss scale $\lambda = 0.05$ with Objaverse Dataset \cite{deitke2023objaverse} rendered by SyncDreamer \cite{liu2023syncdreamer}, filtered by RichDreamer \cite{qiu2023richdreamer} and LGM \cite{tang2024lgm}. For more details, please refer to Appendix~\ref{sec:appendix_implementation_details}.

\subsection{Comparing with Baselines.}
\textbf{Quantitative comparison.}
T3Bench \cite{he2023t} serves as a benchmark for text-to-3D generation, standardizing camera poses for rendering images and utilizing pre-trained models \cite{xu2023imagereward, radford2021learning,li2023blip,achiam2023gpt} to ensure a fair evaluation. 
However, during T3Bench's evaluation, outputs from all other 3D representations are converted to textured meshes, which may cause performance drops. Therefore, we test all methods using their original 3D representations to ensure fairness.

As PlacidDreamer focuses on the generation of individual objects, our evaluations are confined to the single-object part of T3Bench. Considering Fantasia3D \cite{chen2023fantasia3d}, Magic3D \cite{lin2023magic3d}, GeoDream \cite{Ma2023GeoDream}, MVDream \cite{shi2023mvdream}, ProlificDreamer \cite{wang2023prolificdreamer}, and LucidDreamer \cite{EnVision2023luciddreamer} as baseline methods, all of which were tested by T3Bench authors, \citet{he2023t}. The outcomes reveal that PlacidDreamer significantly outperforms the baseline methods in both generative quality and accuracy corresponding to the provided text. Our average quality score surpasses the next-highest ProlificDreamer by 11 points. Since this score is a linear transformation of the ImageReward \cite{xu2023imagereward} score, we annotate the original ImageReward scores, providing a clearer perspective on the superiority of our generative quality.

\noindent\textbf{Qualitative comparison.}
We also conduct qualitative comparisons on several baseline methods which also builds on 3D Gaussian Splatting \cite{kerbl20233d}, including DreamGaussian \cite{tang2023dreamgaussian}, GaussianDreamer \cite{yi2023gaussiandreamer}, GSGEN \cite{chen2023text}, and LucidDreamer \cite{EnVision2023luciddreamer}. In Figure~\ref{fig:baseline}, we present the results of different methods generating the same prompts. DreamGaussian can generate a textured mesh in several minutes, thus, its generation quality is lower compared with other methods. GaussianDreamer and GSGEN leverages end-to-end 3D generation model \cite{jun2023shap, nichol2022point} for initialization with multi-view consistency. However, their final results still have multi-face problems, because the 3D model and Stable Diffusion \cite{rombach2022high} generate towards different directions. LucidDreamer's generations feature bright colors according to the ISM algorithm and exhibit good geometric structure. However, with Latent-Plane enhancing texture and the BSD algorithm, PlacidDreamer achieves superior textures and richer details. 

\subsection{Ablation Studies.}
\label{sec:ablation_study}
We perform ablation studies, as illustrated in Figure~\ref{fig:ablation_study}, and carry out quantitative evaluations using T3Bench, as detailed in Table~\ref{tab:T3Bench}.

\noindent\textbf{Latent-Plane.}
The "Latent-Plane" module in the pipeline serves the functions of initializing 3D Gaussian points and personalizing the diffusion model. We investigate the effects of removing both functionalities (w/o Latent-Plane), removing only initialization (w/o Initialization), and removing only fine-tuning (w/o Personalization) on the experimental results.
Firstly, we remove the whole Latent-Plane module, utilizing the Point-e \cite{nichol2022point} model as the initialization model for 3D Gaussian points, following LucidDreamer. Point-e struggles to comprehend the complex prompts and fails to generate a complete shape for the dog. Additionally, due to the color deviation introduced by Stable Diffusion during generation, the color of the dog's tie is inaccurate. Subsequently, we only remove the initialization function of the Latent-Plane, retaining the fine-tuning part. It can be observed that the Janus problem in the dog's head is resolved. Through extensive experimentation, we find that under the probabilistic supervision of score distillation, 3D Gaussian points do not undergo significant changes in position, making them highly sensitive to initialization. Next, we solely remove the fine-tuning part, keeping the initialization part intact. This results in a more complete structure for the dog's bodies, but issues arise with the Janus problem and tie color. 

\noindent\textbf{BSD.}
We replace BSD with SDS \cite{poole2022dreamfusion} and observe severe over-saturation and some blue artifacts. Compared with BSD, SDS makes 3D Gaussian points harder to converge. To further investigate BSD, we conduct ablation experiments on $\lambda$ to validate its effectiveness in controlling the ratio of $\delta_\mathrm{CG}$ and $\delta_\mathrm{SG}$. It can be observed that when $\lambda=5$, the dog's color is over-saturated. As $\lambda$ increases to the range of 15-25, over-saturation alleviates without significant loss of details. Only when $\lambda$ reaches the value of 35 does the loss of details become apparent. However, at this point, the dog's features are still recognizable. It is worth noting that in the SDS algorithm, CFG can only be set to a large value to get recognizable outcomes.

These results prove that each proposed component significantly contributes to the overall effectiveness of the framework.

\noindent\textbf{Evaluation of BSD in various codebases.}
To further test the robustness and generalization of BSD, we replace other score distillation algorithms in various 3D generation pipelines with BSD. Our results demonstrate that BSD enhances the output quality of each pipeline, as elaborated in Appendix~\ref{sec:appendix_BSD_exp}.

\section{Conclusion}

We have focused on resolving conflicts in current text-to-3D approaches, including conflicts within a single model's guidance and conflicts arising from guidance provided by different models. To address these conflicts, we introduced a novel framework PlacidDreamer with the newly designed Latent-Plane module and the Balanced Score Distillation algorithm to achieve mutual optimization. As a result, we have achieved a more harmonious and balanced text-to-3D generation process, leading to high-fidelity 3D outcomes. We hope our work will inspire further investigations into more harmonious methodologies for 3D generation.


\clearpage
\bibliographystyle{ACM-Reference-Format}
\balance
\bibliography{sample-base}


\begin{thebibliography}{65}


\ifx \showCODEN    \undefined \def \showCODEN     #1{\unskip}     \fi
\ifx \showDOI      \undefined \def \showDOI       #1{#1}\fi
\ifx \showISBNx    \undefined \def \showISBNx     #1{\unskip}     \fi
\ifx \showISBNxiii \undefined \def \showISBNxiii  #1{\unskip}     \fi
\ifx \showISSN     \undefined \def \showISSN      #1{\unskip}     \fi
\ifx \showLCCN     \undefined \def \showLCCN      #1{\unskip}     \fi
\ifx \shownote     \undefined \def \shownote      #1{#1}          \fi
\ifx \showarticletitle \undefined \def \showarticletitle #1{#1}   \fi
\ifx \showURL      \undefined \def \showURL       {\relax}        \fi
\providecommand\bibfield[2]{#2}
\providecommand\bibinfo[2]{#2}
\providecommand\natexlab[1]{#1}
\providecommand\showeprint[2][]{arXiv:#2}

\bibitem[Achiam et~al\mbox{.}(2023)]%
        {achiam2023gpt}
\bibfield{author}{\bibinfo{person}{Josh Achiam}, \bibinfo{person}{Steven Adler}, \bibinfo{person}{Sandhini Agarwal}, \bibinfo{person}{Lama Ahmad}, \bibinfo{person}{Ilge Akkaya}, \bibinfo{person}{Florencia~Leoni Aleman}, \bibinfo{person}{Diogo Almeida}, \bibinfo{person}{Janko Altenschmidt}, \bibinfo{person}{Sam Altman}, \bibinfo{person}{Shyamal Anadkat}, {et~al\mbox{.}}} \bibinfo{year}{2023}\natexlab{}.
\newblock \showarticletitle{Gpt-4 technical report}.
\newblock \bibinfo{journal}{\emph{arXiv preprint arXiv:2303.08774}} (\bibinfo{year}{2023}).
\newblock


\bibitem[Cao et~al\mbox{.}(2023)]%
        {cao2023dreamavatar}
\bibfield{author}{\bibinfo{person}{Yukang Cao}, \bibinfo{person}{Yan-Pei Cao}, \bibinfo{person}{Kai Han}, \bibinfo{person}{Ying Shan}, {and} \bibinfo{person}{Kwan-Yee~K Wong}.} \bibinfo{year}{2023}\natexlab{}.
\newblock \showarticletitle{Dreamavatar: Text-and-shape guided 3d human avatar generation via diffusion models}.
\newblock \bibinfo{journal}{\emph{arXiv preprint arXiv:2304.00916}} (\bibinfo{year}{2023}).
\newblock


\bibitem[Chan et~al\mbox{.}(2022)]%
        {chan2022efficient}
\bibfield{author}{\bibinfo{person}{Eric~R Chan}, \bibinfo{person}{Connor~Z Lin}, \bibinfo{person}{Matthew~A Chan}, \bibinfo{person}{Koki Nagano}, \bibinfo{person}{Boxiao Pan}, \bibinfo{person}{Shalini De~Mello}, \bibinfo{person}{Orazio Gallo}, \bibinfo{person}{Leonidas~J Guibas}, \bibinfo{person}{Jonathan Tremblay}, \bibinfo{person}{Sameh Khamis}, {et~al\mbox{.}}} \bibinfo{year}{2022}\natexlab{}.
\newblock \showarticletitle{Efficient geometry-aware 3d generative adversarial networks}. In \bibinfo{booktitle}{\emph{Proceedings of the IEEE/CVF conference on computer vision and pattern recognition}}. \bibinfo{pages}{16123--16133}.
\newblock


\bibitem[Chen et~al\mbox{.}(2023a)]%
        {chen2023fantasia3d}
\bibfield{author}{\bibinfo{person}{Rui Chen}, \bibinfo{person}{Yongwei Chen}, \bibinfo{person}{Ningxin Jiao}, {and} \bibinfo{person}{Kui Jia}.} \bibinfo{year}{2023}\natexlab{a}.
\newblock \showarticletitle{Fantasia3d: Disentangling geometry and appearance for high-quality text-to-3d content creation}.
\newblock \bibinfo{journal}{\emph{arXiv preprint arXiv:2303.13873}} (\bibinfo{year}{2023}).
\newblock


\bibitem[Chen et~al\mbox{.}(2023c)]%
        {chen2023it3d}
\bibfield{author}{\bibinfo{person}{Yiwen Chen}, \bibinfo{person}{Chi Zhang}, \bibinfo{person}{Xiaofeng Yang}, \bibinfo{person}{Zhongang Cai}, \bibinfo{person}{Gang Yu}, \bibinfo{person}{Lei Yang}, {and} \bibinfo{person}{Guosheng Lin}.} \bibinfo{year}{2023}\natexlab{c}.
\newblock \showarticletitle{It3d: Improved text-to-3d generation with explicit view synthesis}.
\newblock \bibinfo{journal}{\emph{arXiv preprint arXiv:2308.11473}} (\bibinfo{year}{2023}).
\newblock


\bibitem[Chen et~al\mbox{.}(2023b)]%
        {chen2023text}
\bibfield{author}{\bibinfo{person}{Zilong Chen}, \bibinfo{person}{Feng Wang}, {and} \bibinfo{person}{Huaping Liu}.} \bibinfo{year}{2023}\natexlab{b}.
\newblock \showarticletitle{Text-to-3d using gaussian splatting}.
\newblock \bibinfo{journal}{\emph{arXiv preprint arXiv:2309.16585}} (\bibinfo{year}{2023}).
\newblock


\bibitem[Deitke et~al\mbox{.}(2023)]%
        {deitke2023objaverse}
\bibfield{author}{\bibinfo{person}{Matt Deitke}, \bibinfo{person}{Dustin Schwenk}, \bibinfo{person}{Jordi Salvador}, \bibinfo{person}{Luca Weihs}, \bibinfo{person}{Oscar Michel}, \bibinfo{person}{Eli VanderBilt}, \bibinfo{person}{Ludwig Schmidt}, \bibinfo{person}{Kiana Ehsani}, \bibinfo{person}{Aniruddha Kembhavi}, {and} \bibinfo{person}{Ali Farhadi}.} \bibinfo{year}{2023}\natexlab{}.
\newblock \showarticletitle{Objaverse: A universe of annotated 3d objects}. In \bibinfo{booktitle}{\emph{Proceedings of the IEEE/CVF Conference on Computer Vision and Pattern Recognition}}. \bibinfo{pages}{13142--13153}.
\newblock


\bibitem[D{\'e}sid{\'e}ri(2012)]%
        {desideri2012multiple}
\bibfield{author}{\bibinfo{person}{Jean-Antoine D{\'e}sid{\'e}ri}.} \bibinfo{year}{2012}\natexlab{}.
\newblock \showarticletitle{Multiple-gradient descent algorithm (MGDA) for multiobjective optimization}.
\newblock \bibinfo{journal}{\emph{Comptes Rendus Mathematique}} \bibinfo{volume}{350}, \bibinfo{number}{5-6} (\bibinfo{year}{2012}), \bibinfo{pages}{313--318}.
\newblock


\bibitem[Dhariwal and Nichol(2021)]%
        {dhariwal2021diffusion}
\bibfield{author}{\bibinfo{person}{Prafulla Dhariwal} {and} \bibinfo{person}{Alexander Nichol}.} \bibinfo{year}{2021}\natexlab{}.
\newblock \showarticletitle{Diffusion models beat gans on image synthesis}. In \bibinfo{booktitle}{\emph{Proceedings of Advances in Neural Information Processing Systems}}.
\newblock


\bibitem[Guo et~al\mbox{.}(2023)]%
        {threestudio2023}
\bibfield{author}{\bibinfo{person}{Yuan-Chen Guo}, \bibinfo{person}{Ying-Tian Liu}, \bibinfo{person}{Ruizhi Shao}, \bibinfo{person}{Christian Laforte}, \bibinfo{person}{Vikram Voleti}, \bibinfo{person}{Guan Luo}, \bibinfo{person}{Chia-Hao Chen}, \bibinfo{person}{Zi-Xin Zou}, \bibinfo{person}{Chen Wang}, \bibinfo{person}{Yan-Pei Cao}, {and} \bibinfo{person}{Song-Hai Zhang}.} \bibinfo{year}{2023}\natexlab{}.
\newblock \bibinfo{title}{threestudio: A unified framework for 3D content generation}.
\newblock \bibinfo{howpublished}{\url{https://github.com/threestudio-project/threestudio}}.
\newblock


\bibitem[He et~al\mbox{.}(2023)]%
        {he2023t}
\bibfield{author}{\bibinfo{person}{Yuze He}, \bibinfo{person}{Yushi Bai}, \bibinfo{person}{Matthieu Lin}, \bibinfo{person}{Wang Zhao}, \bibinfo{person}{Yubin Hu}, \bibinfo{person}{Jenny Sheng}, \bibinfo{person}{Ran Yi}, \bibinfo{person}{Juanzi Li}, {and} \bibinfo{person}{Yong-Jin Liu}.} \bibinfo{year}{2023}\natexlab{}.
\newblock \bibinfo{title}{T$^3$Bench: Benchmarking Current Progress in Text-to-3D Generation}.
\newblock
\newblock
\showeprint[arxiv]{2310.02977}~[cs.CV]


\bibitem[Hertz et~al\mbox{.}(2023)]%
        {hertz2023delta}
\bibfield{author}{\bibinfo{person}{Amir Hertz}, \bibinfo{person}{Kfir Aberman}, {and} \bibinfo{person}{Daniel Cohen-Or}.} \bibinfo{year}{2023}\natexlab{}.
\newblock \showarticletitle{Delta denoising score}. In \bibinfo{booktitle}{\emph{Proceedings of the IEEE/CVF International Conference on Computer Vision}}. \bibinfo{pages}{2328--2337}.
\newblock


\bibitem[Ho et~al\mbox{.}(2020)]%
        {ho2020denoising}
\bibfield{author}{\bibinfo{person}{Jonathan Ho}, \bibinfo{person}{Ajay Jain}, {and} \bibinfo{person}{Pieter Abbeel}.} \bibinfo{year}{2020}\natexlab{}.
\newblock \showarticletitle{Denoising diffusion probabilistic models}. In \bibinfo{booktitle}{\emph{Proceedings of Advances in Neural Information Processing Systems}}.
\newblock


\bibitem[Hong et~al\mbox{.}(2023)]%
        {hong2023lrm}
\bibfield{author}{\bibinfo{person}{Yicong Hong}, \bibinfo{person}{Kai Zhang}, \bibinfo{person}{Jiuxiang Gu}, \bibinfo{person}{Sai Bi}, \bibinfo{person}{Yang Zhou}, \bibinfo{person}{Difan Liu}, \bibinfo{person}{Feng Liu}, \bibinfo{person}{Kalyan Sunkavalli}, \bibinfo{person}{Trung Bui}, {and} \bibinfo{person}{Hao Tan}.} \bibinfo{year}{2023}\natexlab{}.
\newblock \showarticletitle{Lrm: Large reconstruction model for single image to 3d}.
\newblock \bibinfo{journal}{\emph{arXiv preprint arXiv:2311.04400}} (\bibinfo{year}{2023}).
\newblock


\bibitem[Hu et~al\mbox{.}(2021)]%
        {hu2021lora}
\bibfield{author}{\bibinfo{person}{Edward~J Hu}, \bibinfo{person}{Yelong Shen}, \bibinfo{person}{Phillip Wallis}, \bibinfo{person}{Zeyuan Allen-Zhu}, \bibinfo{person}{Yuanzhi Li}, \bibinfo{person}{Shean Wang}, \bibinfo{person}{Lu Wang}, {and} \bibinfo{person}{Weizhu Chen}.} \bibinfo{year}{2021}\natexlab{}.
\newblock \showarticletitle{Lora: Low-rank adaptation of large language models}.
\newblock \bibinfo{journal}{\emph{arXiv preprint arXiv:2106.09685}} (\bibinfo{year}{2021}).
\newblock


\bibitem[Huang et~al\mbox{.}(2023c)]%
        {huang2023avatarfusion}
\bibfield{author}{\bibinfo{person}{Shuo Huang}, \bibinfo{person}{Zongxin Yang}, \bibinfo{person}{Liangting Li}, \bibinfo{person}{Yi Yang}, {and} \bibinfo{person}{Jia Jia}.} \bibinfo{year}{2023}\natexlab{c}.
\newblock \showarticletitle{AvatarFusion: Zero-shot Generation of Clothing-Decoupled 3D Avatars Using 2D Diffusion}. In \bibinfo{booktitle}{\emph{Proceedings of the 31st ACM International Conference on Multimedia}}. \bibinfo{pages}{5734--5745}.
\newblock


\bibitem[Huang et~al\mbox{.}(2023a)]%
        {huang2023dreamtime}
\bibfield{author}{\bibinfo{person}{Yukun Huang}, \bibinfo{person}{Jianan Wang}, \bibinfo{person}{Yukai Shi}, \bibinfo{person}{Xianbiao Qi}, \bibinfo{person}{Zheng-Jun Zha}, {and} \bibinfo{person}{Lei Zhang}.} \bibinfo{year}{2023}\natexlab{a}.
\newblock \showarticletitle{DreamTime: An Improved Optimization Strategy for Text-to-3D Content Creation}.
\newblock \bibinfo{journal}{\emph{arXiv preprint arXiv:2306.12422}} (\bibinfo{year}{2023}).
\newblock


\bibitem[Huang et~al\mbox{.}(2023b)]%
        {huang2023dreamwaltz}
\bibfield{author}{\bibinfo{person}{Yukun Huang}, \bibinfo{person}{Jianan Wang}, \bibinfo{person}{Ailing Zeng}, \bibinfo{person}{He Cao}, \bibinfo{person}{Xianbiao Qi}, \bibinfo{person}{Yukai Shi}, \bibinfo{person}{Zheng-Jun Zha}, {and} \bibinfo{person}{Lei Zhang}.} \bibinfo{year}{2023}\natexlab{b}.
\newblock \showarticletitle{DreamWaltz: Make a Scene with Complex 3D Animatable Avatars}.
\newblock \bibinfo{journal}{\emph{arXiv preprint arXiv:2305.12529}} (\bibinfo{year}{2023}).
\newblock


\bibitem[Jain et~al\mbox{.}(2022)]%
        {jain2022zero}
\bibfield{author}{\bibinfo{person}{Ajay Jain}, \bibinfo{person}{Ben Mildenhall}, \bibinfo{person}{Jonathan~T Barron}, \bibinfo{person}{Pieter Abbeel}, {and} \bibinfo{person}{Ben Poole}.} \bibinfo{year}{2022}\natexlab{}.
\newblock \showarticletitle{Zero-shot text-guided object generation with dream fields}. In \bibinfo{booktitle}{\emph{Proceedings of the IEEE/CVF Conference on Computer Vision and Pattern Recognition}}. \bibinfo{pages}{867--876}.
\newblock


\bibitem[Jiang et~al\mbox{.}(2023)]%
        {jiang2023avatarcraft}
\bibfield{author}{\bibinfo{person}{Ruixiang Jiang}, \bibinfo{person}{Can Wang}, \bibinfo{person}{Jingbo Zhang}, \bibinfo{person}{Menglei Chai}, \bibinfo{person}{Mingming He}, \bibinfo{person}{Dongdong Chen}, {and} \bibinfo{person}{Jing Liao}.} \bibinfo{year}{2023}\natexlab{}.
\newblock \showarticletitle{AvatarCraft: Transforming Text into Neural Human Avatars with Parameterized Shape and Pose Control}.
\newblock \bibinfo{journal}{\emph{arXiv preprint arXiv:2303.17606}} (\bibinfo{year}{2023}).
\newblock


\bibitem[Jun and Nichol(2023)]%
        {jun2023shap}
\bibfield{author}{\bibinfo{person}{Heewoo Jun} {and} \bibinfo{person}{Alex Nichol}.} \bibinfo{year}{2023}\natexlab{}.
\newblock \showarticletitle{Shap-e: Generating conditional 3d implicit functions}.
\newblock \bibinfo{journal}{\emph{arXiv preprint arXiv:2305.02463}} (\bibinfo{year}{2023}).
\newblock


\bibitem[Katzir et~al\mbox{.}(2023)]%
        {katzir2023noise}
\bibfield{author}{\bibinfo{person}{Oren Katzir}, \bibinfo{person}{Or Patashnik}, \bibinfo{person}{Daniel Cohen-Or}, {and} \bibinfo{person}{Dani Lischinski}.} \bibinfo{year}{2023}\natexlab{}.
\newblock \showarticletitle{Noise-free score distillation}.
\newblock \bibinfo{journal}{\emph{arXiv preprint arXiv:2310.17590}} (\bibinfo{year}{2023}).
\newblock


\bibitem[Kerbl et~al\mbox{.}(2023)]%
        {kerbl20233d}
\bibfield{author}{\bibinfo{person}{Bernhard Kerbl}, \bibinfo{person}{Georgios Kopanas}, \bibinfo{person}{Thomas Leimk{\"u}hler}, {and} \bibinfo{person}{George Drettakis}.} \bibinfo{year}{2023}\natexlab{}.
\newblock \showarticletitle{3D Gaussian Splatting for Real-Time Radiance Field Rendering}.
\newblock \bibinfo{journal}{\emph{ACM Transactions on Graphics}} \bibinfo{volume}{42}, \bibinfo{number}{4} (\bibinfo{year}{2023}).
\newblock


\bibitem[Li et~al\mbox{.}(2023b)]%
        {li2023blip}
\bibfield{author}{\bibinfo{person}{Junnan Li}, \bibinfo{person}{Dongxu Li}, \bibinfo{person}{Silvio Savarese}, {and} \bibinfo{person}{Steven Hoi}.} \bibinfo{year}{2023}\natexlab{b}.
\newblock \showarticletitle{Blip-2: Bootstrapping language-image pre-training with frozen image encoders and large language models}.
\newblock \bibinfo{journal}{\emph{arXiv preprint arXiv:2301.12597}} (\bibinfo{year}{2023}).
\newblock


\bibitem[Li et~al\mbox{.}(2023a)]%
        {li2023sweetdreamer}
\bibfield{author}{\bibinfo{person}{Weiyu Li}, \bibinfo{person}{Rui Chen}, \bibinfo{person}{Xuelin Chen}, {and} \bibinfo{person}{Ping Tan}.} \bibinfo{year}{2023}\natexlab{a}.
\newblock \showarticletitle{Sweetdreamer: Aligning geometric priors in 2d diffusion for consistent text-to-3d}.
\newblock \bibinfo{journal}{\emph{arXiv preprint arXiv:2310.02596}} (\bibinfo{year}{2023}).
\newblock


\bibitem[Liang et~al\mbox{.}(2023)]%
        {EnVision2023luciddreamer}
\bibfield{author}{\bibinfo{person}{Yixun Liang}, \bibinfo{person}{Xin Yang}, \bibinfo{person}{Jiantao Lin}, \bibinfo{person}{Haodong Li}, \bibinfo{person}{Xiaogang Xu}, {and} \bibinfo{person}{Yingcong Chen}.} \bibinfo{year}{2023}\natexlab{}.
\newblock \bibinfo{title}{LucidDreamer: Towards High-Fidelity Text-to-3D Generation via Interval Score Matching}.
\newblock
\newblock
\showeprint[arxiv]{2311.11284}~[cs.CV]


\bibitem[Lin et~al\mbox{.}(2023a)]%
        {lin2023magic3d}
\bibfield{author}{\bibinfo{person}{Chen-Hsuan Lin}, \bibinfo{person}{Jun Gao}, \bibinfo{person}{Luming Tang}, \bibinfo{person}{Towaki Takikawa}, \bibinfo{person}{Xiaohui Zeng}, \bibinfo{person}{Xun Huang}, \bibinfo{person}{Karsten Kreis}, \bibinfo{person}{Sanja Fidler}, \bibinfo{person}{Ming-Yu Liu}, {and} \bibinfo{person}{Tsung-Yi Lin}.} \bibinfo{year}{2023}\natexlab{a}.
\newblock \showarticletitle{Magic3d: High-resolution text-to-3d content creation}. In \bibinfo{booktitle}{\emph{Proceedings of the IEEE/CVF Conference on Computer Vision and Pattern Recognition}}. \bibinfo{pages}{300--309}.
\newblock


\bibitem[Lin et~al\mbox{.}(2023b)]%
        {lin2023consistent123}
\bibfield{author}{\bibinfo{person}{Yukang Lin}, \bibinfo{person}{Haonan Han}, \bibinfo{person}{Chaoqun Gong}, \bibinfo{person}{Zunnan Xu}, \bibinfo{person}{Yachao Zhang}, {and} \bibinfo{person}{Xiu Li}.} \bibinfo{year}{2023}\natexlab{b}.
\newblock \showarticletitle{Consistent123: One image to highly consistent 3d asset using case-aware diffusion priors}.
\newblock \bibinfo{journal}{\emph{arXiv preprint arXiv:2309.17261}} (\bibinfo{year}{2023}).
\newblock


\bibitem[Liu et~al\mbox{.}(2023b)]%
        {liu2023one}
\bibfield{author}{\bibinfo{person}{Minghua Liu}, \bibinfo{person}{Ruoxi Shi}, \bibinfo{person}{Linghao Chen}, \bibinfo{person}{Zhuoyang Zhang}, \bibinfo{person}{Chao Xu}, \bibinfo{person}{Xinyue Wei}, \bibinfo{person}{Hansheng Chen}, \bibinfo{person}{Chong Zeng}, \bibinfo{person}{Jiayuan Gu}, {and} \bibinfo{person}{Hao Su}.} \bibinfo{year}{2023}\natexlab{b}.
\newblock \showarticletitle{One-2-3-45++: Fast single image to 3d objects with consistent multi-view generation and 3d diffusion}.
\newblock \bibinfo{journal}{\emph{arXiv preprint arXiv:2311.07885}} (\bibinfo{year}{2023}).
\newblock


\bibitem[Liu et~al\mbox{.}(2023c)]%
        {liu2023zero}
\bibfield{author}{\bibinfo{person}{Ruoshi Liu}, \bibinfo{person}{Rundi Wu}, \bibinfo{person}{Basile Van~Hoorick}, \bibinfo{person}{Pavel Tokmakov}, \bibinfo{person}{Sergey Zakharov}, {and} \bibinfo{person}{Carl Vondrick}.} \bibinfo{year}{2023}\natexlab{c}.
\newblock \showarticletitle{Zero-1-to-3: Zero-shot one image to 3d object}. In \bibinfo{booktitle}{\emph{Proceedings of the IEEE/CVF International Conference on Computer Vision}}. \bibinfo{pages}{9298--9309}.
\newblock


\bibitem[Liu et~al\mbox{.}(2023a)]%
        {liu2023syncdreamer}
\bibfield{author}{\bibinfo{person}{Yuan Liu}, \bibinfo{person}{Cheng Lin}, \bibinfo{person}{Zijiao Zeng}, \bibinfo{person}{Xiaoxiao Long}, \bibinfo{person}{Lingjie Liu}, \bibinfo{person}{Taku Komura}, {and} \bibinfo{person}{Wenping Wang}.} \bibinfo{year}{2023}\natexlab{a}.
\newblock \showarticletitle{SyncDreamer: Generating Multiview-consistent Images from a Single-view Image}.
\newblock \bibinfo{journal}{\emph{arXiv preprint arXiv:2309.03453}} (\bibinfo{year}{2023}).
\newblock


\bibitem[Long et~al\mbox{.}(2023)]%
        {long2023wonder3d}
\bibfield{author}{\bibinfo{person}{Xiaoxiao Long}, \bibinfo{person}{Yuan-Chen Guo}, \bibinfo{person}{Cheng Lin}, \bibinfo{person}{Yuan Liu}, \bibinfo{person}{Zhiyang Dou}, \bibinfo{person}{Lingjie Liu}, \bibinfo{person}{Yuexin Ma}, \bibinfo{person}{Song-Hai Zhang}, \bibinfo{person}{Marc Habermann}, \bibinfo{person}{Christian Theobalt}, {et~al\mbox{.}}} \bibinfo{year}{2023}\natexlab{}.
\newblock \showarticletitle{Wonder3d: Single image to 3d using cross-domain diffusion}.
\newblock \bibinfo{journal}{\emph{arXiv preprint arXiv:2310.15008}} (\bibinfo{year}{2023}).
\newblock


\bibitem[Lorraine et~al\mbox{.}(2023)]%
        {lorraine2023att3d}
\bibfield{author}{\bibinfo{person}{Jonathan Lorraine}, \bibinfo{person}{Kevin Xie}, \bibinfo{person}{Xiaohui Zeng}, \bibinfo{person}{Chen-Hsuan Lin}, \bibinfo{person}{Towaki Takikawa}, \bibinfo{person}{Nicholas Sharp}, \bibinfo{person}{Tsung-Yi Lin}, \bibinfo{person}{Ming-Yu Liu}, \bibinfo{person}{Sanja Fidler}, {and} \bibinfo{person}{James Lucas}.} \bibinfo{year}{2023}\natexlab{}.
\newblock \showarticletitle{Att3d: Amortized text-to-3d object synthesis}. In \bibinfo{booktitle}{\emph{Proceedings of the IEEE/CVF International Conference on Computer Vision}}. \bibinfo{pages}{17946--17956}.
\newblock


\bibitem[Ma et~al\mbox{.}(2023)]%
        {Ma2023GeoDream}
\bibfield{author}{\bibinfo{person}{Baorui Ma}, \bibinfo{person}{Haoge Deng}, \bibinfo{person}{Junsheng Zhou}, \bibinfo{person}{Yu-Shen Liu}, \bibinfo{person}{Tiejun Huang}, {and} \bibinfo{person}{Xinlong Wang}.} \bibinfo{year}{2023}\natexlab{}.
\newblock \showarticletitle{GeoDream: Disentangling 2D and Geometric Priors for High-Fidelity and Consistent 3D Generation}.
\newblock \bibinfo{journal}{\emph{arXiv preprint arXiv:2311.17971}}.
\newblock


\bibitem[Metzer et~al\mbox{.}(2023)]%
        {metzer2023latent}
\bibfield{author}{\bibinfo{person}{Gal Metzer}, \bibinfo{person}{Elad Richardson}, \bibinfo{person}{Or Patashnik}, \bibinfo{person}{Raja Giryes}, {and} \bibinfo{person}{Daniel Cohen-Or}.} \bibinfo{year}{2023}\natexlab{}.
\newblock \showarticletitle{Latent-nerf for shape-guided generation of 3d shapes and textures}. In \bibinfo{booktitle}{\emph{Proceedings of the IEEE/CVF Conference on Computer Vision and Pattern Recognition}}. \bibinfo{pages}{12663--12673}.
\newblock


\bibitem[Mildenhall et~al\mbox{.}(2021)]%
        {mildenhall2021nerf}
\bibfield{author}{\bibinfo{person}{Ben Mildenhall}, \bibinfo{person}{Pratul~P Srinivasan}, \bibinfo{person}{Matthew Tancik}, \bibinfo{person}{Jonathan~T Barron}, \bibinfo{person}{Ravi Ramamoorthi}, {and} \bibinfo{person}{Ren Ng}.} \bibinfo{year}{2021}\natexlab{}.
\newblock \showarticletitle{Nerf: Representing scenes as neural radiance fields for view synthesis}.
\newblock \bibinfo{journal}{\emph{Commun. ACM}} \bibinfo{volume}{65}, \bibinfo{number}{1} (\bibinfo{year}{2021}), \bibinfo{pages}{99--106}.
\newblock


\bibitem[Nichol et~al\mbox{.}(2022)]%
        {nichol2022point}
\bibfield{author}{\bibinfo{person}{Alex Nichol}, \bibinfo{person}{Heewoo Jun}, \bibinfo{person}{Prafulla Dhariwal}, \bibinfo{person}{Pamela Mishkin}, {and} \bibinfo{person}{Mark Chen}.} \bibinfo{year}{2022}\natexlab{}.
\newblock \showarticletitle{Point-e: A system for generating 3d point clouds from complex prompts}.
\newblock \bibinfo{journal}{\emph{arXiv preprint arXiv:2212.08751}} (\bibinfo{year}{2022}).
\newblock


\bibitem[Nichol and Dhariwal(2021)]%
        {nichol2021improved}
\bibfield{author}{\bibinfo{person}{Alexander~Quinn Nichol} {and} \bibinfo{person}{Prafulla Dhariwal}.} \bibinfo{year}{2021}\natexlab{}.
\newblock \showarticletitle{Improved denoising diffusion probabilistic models}. In \bibinfo{booktitle}{\emph{International Conference on Machine Learning}}. PMLR, \bibinfo{pages}{8162--8171}.
\newblock


\bibitem[Poole et~al\mbox{.}(2022)]%
        {poole2022dreamfusion}
\bibfield{author}{\bibinfo{person}{Ben Poole}, \bibinfo{person}{Ajay Jain}, \bibinfo{person}{Jonathan~T Barron}, {and} \bibinfo{person}{Ben Mildenhall}.} \bibinfo{year}{2022}\natexlab{}.
\newblock \showarticletitle{Dreamfusion: Text-to-3d using 2d diffusion}.
\newblock \bibinfo{journal}{\emph{arXiv preprint arXiv:2209.14988}} (\bibinfo{year}{2022}).
\newblock


\bibitem[Qian et~al\mbox{.}(2023)]%
        {qian2023magic123}
\bibfield{author}{\bibinfo{person}{Guocheng Qian}, \bibinfo{person}{Jinjie Mai}, \bibinfo{person}{Abdullah Hamdi}, \bibinfo{person}{Jian Ren}, \bibinfo{person}{Aliaksandr Siarohin}, \bibinfo{person}{Bing Li}, \bibinfo{person}{Hsin-Ying Lee}, \bibinfo{person}{Ivan Skorokhodov}, \bibinfo{person}{Peter Wonka}, \bibinfo{person}{Sergey Tulyakov}, {et~al\mbox{.}}} \bibinfo{year}{2023}\natexlab{}.
\newblock \showarticletitle{Magic123: One image to high-quality 3d object generation using both 2d and 3d diffusion priors}.
\newblock \bibinfo{journal}{\emph{arXiv preprint arXiv:2306.17843}} (\bibinfo{year}{2023}).
\newblock


\bibitem[Qiu et~al\mbox{.}(2023)]%
        {qiu2023richdreamer}
\bibfield{author}{\bibinfo{person}{Lingteng Qiu}, \bibinfo{person}{Guanying Chen}, \bibinfo{person}{Xiaodong Gu}, \bibinfo{person}{Qi zuo}, \bibinfo{person}{Mutian Xu}, \bibinfo{person}{Yushuang Wu}, \bibinfo{person}{Weihao Yuan}, \bibinfo{person}{Zilong Dong}, \bibinfo{person}{Liefeng Bo}, {and} \bibinfo{person}{Xiaoguang Han}.} \bibinfo{year}{2023}\natexlab{}.
\newblock \showarticletitle{RichDreamer: A Generalizable Normal-Depth Diffusion Model for Detail Richness in Text-to-3D}.
\newblock \bibinfo{journal}{\emph{arXiv preprint arXiv:2311.16918}} (\bibinfo{year}{2023}).
\newblock


\bibitem[Radford et~al\mbox{.}(2021)]%
        {radford2021learning}
\bibfield{author}{\bibinfo{person}{Alec Radford}, \bibinfo{person}{Jong~Wook Kim}, \bibinfo{person}{Chris Hallacy}, \bibinfo{person}{Aditya Ramesh}, \bibinfo{person}{Gabriel Goh}, \bibinfo{person}{Sandhini Agarwal}, \bibinfo{person}{Girish Sastry}, \bibinfo{person}{Amanda Askell}, \bibinfo{person}{Pamela Mishkin}, \bibinfo{person}{Jack Clark}, {et~al\mbox{.}}} \bibinfo{year}{2021}\natexlab{}.
\newblock \showarticletitle{Learning transferable visual models from natural language supervision}. In \bibinfo{booktitle}{\emph{International conference on machine learning}}. PMLR, \bibinfo{pages}{8748--8763}.
\newblock


\bibitem[Rombach et~al\mbox{.}(2022)]%
        {rombach2022high}
\bibfield{author}{\bibinfo{person}{Robin Rombach}, \bibinfo{person}{Andreas Blattmann}, \bibinfo{person}{Dominik Lorenz}, \bibinfo{person}{Patrick Esser}, {and} \bibinfo{person}{Bj{\"o}rn Ommer}.} \bibinfo{year}{2022}\natexlab{}.
\newblock \showarticletitle{High-resolution image synthesis with latent diffusion models}. In \bibinfo{booktitle}{\emph{Proceedings of the IEEE/CVF conference on computer vision and pattern recognition}}. \bibinfo{pages}{10684--10695}.
\newblock


\bibitem[Ronneberger et~al\mbox{.}(2015)]%
        {ronneberger2015u}
\bibfield{author}{\bibinfo{person}{Olaf Ronneberger}, \bibinfo{person}{Philipp Fischer}, {and} \bibinfo{person}{Thomas Brox}.} \bibinfo{year}{2015}\natexlab{}.
\newblock \showarticletitle{U-net: Convolutional networks for biomedical image segmentation}. In \bibinfo{booktitle}{\emph{Medical image computing and computer-assisted intervention--MICCAI 2015: 18th international conference, Munich, Germany, October 5-9, 2015, proceedings, part III 18}}. Springer, \bibinfo{pages}{234--241}.
\newblock


\bibitem[Sargent et~al\mbox{.}(2023)]%
        {sargent2023zeronvs}
\bibfield{author}{\bibinfo{person}{Kyle Sargent}, \bibinfo{person}{Zizhang Li}, \bibinfo{person}{Tanmay Shah}, \bibinfo{person}{Charles Herrmann}, \bibinfo{person}{Hong-Xing Yu}, \bibinfo{person}{Yunzhi Zhang}, \bibinfo{person}{Eric~Ryan Chan}, \bibinfo{person}{Dmitry Lagun}, \bibinfo{person}{Li Fei-Fei}, \bibinfo{person}{Deqing Sun}, {et~al\mbox{.}}} \bibinfo{year}{2023}\natexlab{}.
\newblock \showarticletitle{Zeronvs: Zero-shot 360-degree view synthesis from a single real image}.
\newblock \bibinfo{journal}{\emph{arXiv preprint arXiv:2310.17994}} (\bibinfo{year}{2023}).
\newblock


\bibitem[Shen et~al\mbox{.}(2021)]%
        {shen2021deep}
\bibfield{author}{\bibinfo{person}{Tianchang Shen}, \bibinfo{person}{Jun Gao}, \bibinfo{person}{Kangxue Yin}, \bibinfo{person}{Ming-Yu Liu}, {and} \bibinfo{person}{Sanja Fidler}.} \bibinfo{year}{2021}\natexlab{}.
\newblock \showarticletitle{Deep marching tetrahedra: a hybrid representation for high-resolution 3d shape synthesis}.
\newblock \bibinfo{journal}{\emph{Advances in Neural Information Processing Systems}}  \bibinfo{volume}{34} (\bibinfo{year}{2021}), \bibinfo{pages}{6087--6101}.
\newblock


\bibitem[Shi et~al\mbox{.}(2023)]%
        {shi2023mvdream}
\bibfield{author}{\bibinfo{person}{Yichun Shi}, \bibinfo{person}{Peng Wang}, \bibinfo{person}{Jianglong Ye}, \bibinfo{person}{Mai Long}, \bibinfo{person}{Kejie Li}, {and} \bibinfo{person}{Xiao Yang}.} \bibinfo{year}{2023}\natexlab{}.
\newblock \showarticletitle{Mvdream: Multi-view diffusion for 3d generation}.
\newblock \bibinfo{journal}{\emph{arXiv preprint arXiv:2308.16512}} (\bibinfo{year}{2023}).
\newblock


\bibitem[Sohl-Dickstein et~al\mbox{.}(2015)]%
        {sohl-dickstein15}
\bibfield{author}{\bibinfo{person}{Jascha Sohl-Dickstein}, \bibinfo{person}{Eric Weiss}, \bibinfo{person}{Niru Maheswaranathan}, {and} \bibinfo{person}{Surya Ganguli}.} \bibinfo{year}{2015}\natexlab{}.
\newblock \showarticletitle{Deep Unsupervised Learning using Nonequilibrium Thermodynamics}. In \bibinfo{booktitle}{\emph{Proceedings of International Conference on Machine Learning}}.
\newblock


\bibitem[Song and Ermon(2019)]%
        {song2019generative}
\bibfield{author}{\bibinfo{person}{Yang Song} {and} \bibinfo{person}{Stefano Ermon}.} \bibinfo{year}{2019}\natexlab{}.
\newblock \showarticletitle{Generative modeling by estimating gradients of the data distribution}.
\newblock \bibinfo{journal}{\emph{Advances in neural information processing systems}}  \bibinfo{volume}{32} (\bibinfo{year}{2019}).
\newblock


\bibitem[Sun et~al\mbox{.}(2023)]%
        {sun2023dreamcraft3d}
\bibfield{author}{\bibinfo{person}{Jingxiang Sun}, \bibinfo{person}{Bo Zhang}, \bibinfo{person}{Ruizhi Shao}, \bibinfo{person}{Lizhen Wang}, \bibinfo{person}{Wen Liu}, \bibinfo{person}{Zhenda Xie}, {and} \bibinfo{person}{Yebin Liu}.} \bibinfo{year}{2023}\natexlab{}.
\newblock \showarticletitle{Dreamcraft3d: Hierarchical 3d generation with bootstrapped diffusion prior}.
\newblock \bibinfo{journal}{\emph{arXiv preprint arXiv:2310.16818}} (\bibinfo{year}{2023}).
\newblock


\bibitem[Tang et~al\mbox{.}(2023b)]%
        {tang2023stable}
\bibfield{author}{\bibinfo{person}{Boshi Tang}, \bibinfo{person}{Jianan Wang}, \bibinfo{person}{Zhiyong Wu}, {and} \bibinfo{person}{Lei Zhang}.} \bibinfo{year}{2023}\natexlab{b}.
\newblock \showarticletitle{Stable Score Distillation for High-Quality 3D Generation}.
\newblock \bibinfo{journal}{\emph{arXiv preprint arXiv:2312.09305}} (\bibinfo{year}{2023}).
\newblock


\bibitem[Tang et~al\mbox{.}(2024)]%
        {tang2024lgm}
\bibfield{author}{\bibinfo{person}{Jiaxiang Tang}, \bibinfo{person}{Zhaoxi Chen}, \bibinfo{person}{Xiaokang Chen}, \bibinfo{person}{Tengfei Wang}, \bibinfo{person}{Gang Zeng}, {and} \bibinfo{person}{Ziwei Liu}.} \bibinfo{year}{2024}\natexlab{}.
\newblock \showarticletitle{LGM: Large Multi-View Gaussian Model for High-Resolution 3D Content Creation}.
\newblock \bibinfo{journal}{\emph{arXiv preprint arXiv:2402.05054}} (\bibinfo{year}{2024}).
\newblock


\bibitem[Tang et~al\mbox{.}(2023a)]%
        {tang2023dreamgaussian}
\bibfield{author}{\bibinfo{person}{Jiaxiang Tang}, \bibinfo{person}{Jiawei Ren}, \bibinfo{person}{Hang Zhou}, \bibinfo{person}{Ziwei Liu}, {and} \bibinfo{person}{Gang Zeng}.} \bibinfo{year}{2023}\natexlab{a}.
\newblock \showarticletitle{Dreamgaussian: Generative gaussian splatting for efficient 3d content creation}.
\newblock \bibinfo{journal}{\emph{arXiv preprint arXiv:2309.16653}} (\bibinfo{year}{2023}).
\newblock


\bibitem[Wang et~al\mbox{.}(2023a)]%
        {wang2023score}
\bibfield{author}{\bibinfo{person}{Haochen Wang}, \bibinfo{person}{Xiaodan Du}, \bibinfo{person}{Jiahao Li}, \bibinfo{person}{Raymond~A Yeh}, {and} \bibinfo{person}{Greg Shakhnarovich}.} \bibinfo{year}{2023}\natexlab{a}.
\newblock \showarticletitle{Score jacobian chaining: Lifting pretrained 2d diffusion models for 3d generation}. In \bibinfo{booktitle}{\emph{Proceedings of the IEEE/CVF Conference on Computer Vision and Pattern Recognition}}. \bibinfo{pages}{12619--12629}.
\newblock


\bibitem[Wang et~al\mbox{.}(2023b)]%
        {wang2023steindreamer}
\bibfield{author}{\bibinfo{person}{Peihao Wang}, \bibinfo{person}{Zhiwen Fan}, \bibinfo{person}{Dejia Xu}, \bibinfo{person}{Dilin Wang}, \bibinfo{person}{Sreyas Mohan}, \bibinfo{person}{Forrest Iandola}, \bibinfo{person}{Rakesh Ranjan}, \bibinfo{person}{Yilei Li}, \bibinfo{person}{Qiang Liu}, \bibinfo{person}{Zhangyang Wang}, {et~al\mbox{.}}} \bibinfo{year}{2023}\natexlab{b}.
\newblock \showarticletitle{SteinDreamer: Variance Reduction for Text-to-3D Score Distillation via Stein Identity}.
\newblock \bibinfo{journal}{\emph{arXiv preprint arXiv:2401.00604}} (\bibinfo{year}{2023}).
\newblock


\bibitem[Wang et~al\mbox{.}(2023c)]%
        {wang2023prolificdreamer}
\bibfield{author}{\bibinfo{person}{Zhengyi Wang}, \bibinfo{person}{Cheng Lu}, \bibinfo{person}{Yikai Wang}, \bibinfo{person}{Fan Bao}, \bibinfo{person}{Chongxuan Li}, \bibinfo{person}{Hang Su}, {and} \bibinfo{person}{Jun Zhu}.} \bibinfo{year}{2023}\natexlab{c}.
\newblock \showarticletitle{ProlificDreamer: High-Fidelity and Diverse Text-to-3D Generation with Variational Score Distillation}.
\newblock \bibinfo{journal}{\emph{arXiv preprint arXiv:2305.16213}} (\bibinfo{year}{2023}).
\newblock


\bibitem[Wu et~al\mbox{.}(2024)]%
        {wu2024hd}
\bibfield{author}{\bibinfo{person}{Jinbo Wu}, \bibinfo{person}{Xiaobo Gao}, \bibinfo{person}{Xing Liu}, \bibinfo{person}{Zhengyang Shen}, \bibinfo{person}{Chen Zhao}, \bibinfo{person}{Haocheng Feng}, \bibinfo{person}{Jingtuo Liu}, {and} \bibinfo{person}{Errui Ding}.} \bibinfo{year}{2024}\natexlab{}.
\newblock \showarticletitle{Hd-fusion: Detailed text-to-3d generation leveraging multiple noise estimation}. In \bibinfo{booktitle}{\emph{Proceedings of the IEEE/CVF Winter Conference on Applications of Computer Vision}}. \bibinfo{pages}{3202--3211}.
\newblock


\bibitem[Xu et~al\mbox{.}(2023a)]%
        {xu2023imagereward}
\bibfield{author}{\bibinfo{person}{Jiazheng Xu}, \bibinfo{person}{Xiao Liu}, \bibinfo{person}{Yuchen Wu}, \bibinfo{person}{Yuxuan Tong}, \bibinfo{person}{Qinkai Li}, \bibinfo{person}{Ming Ding}, \bibinfo{person}{Jie Tang}, {and} \bibinfo{person}{Yuxiao Dong}.} \bibinfo{year}{2023}\natexlab{a}.
\newblock \showarticletitle{Imagereward: Learning and evaluating human preferences for text-to-image generation}.
\newblock \bibinfo{journal}{\emph{arXiv preprint arXiv:2304.05977}} (\bibinfo{year}{2023}).
\newblock


\bibitem[Xu et~al\mbox{.}(2023b)]%
        {xu2023seeavatar}
\bibfield{author}{\bibinfo{person}{Yuanyou Xu}, \bibinfo{person}{Zongxin Yang}, {and} \bibinfo{person}{Yi Yang}.} \bibinfo{year}{2023}\natexlab{b}.
\newblock \showarticletitle{SEEAvatar: Photorealistic Text-to-3D Avatar Generation with Constrained Geometry and Appearance}.
\newblock \bibinfo{journal}{\emph{arXiv preprint arXiv:2312.08889}} (\bibinfo{year}{2023}).
\newblock


\bibitem[Yang et~al\mbox{.}(2023)]%
        {yang2023consistnet}
\bibfield{author}{\bibinfo{person}{Jiayu Yang}, \bibinfo{person}{Ziang Cheng}, \bibinfo{person}{Yunfei Duan}, \bibinfo{person}{Pan Ji}, {and} \bibinfo{person}{Hongdong Li}.} \bibinfo{year}{2023}\natexlab{}.
\newblock \showarticletitle{Consistnet: Enforcing 3d consistency for multi-view images diffusion}.
\newblock \bibinfo{journal}{\emph{arXiv preprint arXiv:2310.10343}} (\bibinfo{year}{2023}).
\newblock


\bibitem[Yi et~al\mbox{.}(2023)]%
        {yi2023gaussiandreamer}
\bibfield{author}{\bibinfo{person}{Taoran Yi}, \bibinfo{person}{Jiemin Fang}, \bibinfo{person}{Guanjun Wu}, \bibinfo{person}{Lingxi Xie}, \bibinfo{person}{Xiaopeng Zhang}, \bibinfo{person}{Wenyu Liu}, \bibinfo{person}{Qi Tian}, {and} \bibinfo{person}{Xinggang Wang}.} \bibinfo{year}{2023}\natexlab{}.
\newblock \showarticletitle{Gaussiandreamer: Fast generation from text to 3d gaussian splatting with point cloud priors}.
\newblock \bibinfo{journal}{\emph{arXiv preprint arXiv:2310.08529}} (\bibinfo{year}{2023}).
\newblock


\bibitem[Yu et~al\mbox{.}(2023)]%
        {yu2023text}
\bibfield{author}{\bibinfo{person}{Xin Yu}, \bibinfo{person}{Yuan-Chen Guo}, \bibinfo{person}{Yangguang Li}, \bibinfo{person}{Ding Liang}, \bibinfo{person}{Song-Hai Zhang}, {and} \bibinfo{person}{Xiaojuan Qi}.} \bibinfo{year}{2023}\natexlab{}.
\newblock \showarticletitle{Text-to-3d with classifier score distillation}.
\newblock \bibinfo{journal}{\emph{arXiv preprint arXiv:2310.19415}} (\bibinfo{year}{2023}).
\newblock


\bibitem[Zeng et~al\mbox{.}(2023)]%
        {zeng2023avatarbooth}
\bibfield{author}{\bibinfo{person}{Yifei Zeng}, \bibinfo{person}{Yuanxun Lu}, \bibinfo{person}{Xinya Ji}, \bibinfo{person}{Yao Yao}, \bibinfo{person}{Hao Zhu}, {and} \bibinfo{person}{Xun Cao}.} \bibinfo{year}{2023}\natexlab{}.
\newblock \showarticletitle{AvatarBooth: High-Quality and Customizable 3D Human Avatar Generation}.
\newblock \bibinfo{journal}{\emph{arXiv preprint arXiv:2306.09864}} (\bibinfo{year}{2023}).
\newblock


\bibitem[Zhao et~al\mbox{.}(2023)]%
        {zhao2023efficientdreamer}
\bibfield{author}{\bibinfo{person}{Minda Zhao}, \bibinfo{person}{Chaoyi Zhao}, \bibinfo{person}{Xinyue Liang}, \bibinfo{person}{Lincheng Li}, \bibinfo{person}{Zeng Zhao}, \bibinfo{person}{Zhipeng Hu}, \bibinfo{person}{Changjie Fan}, {and} \bibinfo{person}{Xin Yu}.} \bibinfo{year}{2023}\natexlab{}.
\newblock \showarticletitle{EfficientDreamer: High-Fidelity and Robust 3D Creation via Orthogonal-view Diffusion Prior}.
\newblock \bibinfo{journal}{\emph{arXiv preprint arXiv:2308.13223}} (\bibinfo{year}{2023}).
\newblock


\bibitem[Zhu and Zhuang(2023)]%
        {zhu2023hifa}
\bibfield{author}{\bibinfo{person}{Joseph Zhu} {and} \bibinfo{person}{Peiye Zhuang}.} \bibinfo{year}{2023}\natexlab{}.
\newblock \showarticletitle{HiFA: High-fidelity Text-to-3D with Advanced Diffusion Guidance}.
\newblock \bibinfo{journal}{\emph{arXiv preprint arXiv:2305.18766}} (\bibinfo{year}{2023}).
\newblock


\end{thebibliography}
\clearpage
\appendix
\begin{figure*}[h]
\begin{center}
\centerline{\includegraphics[width=\textwidth]{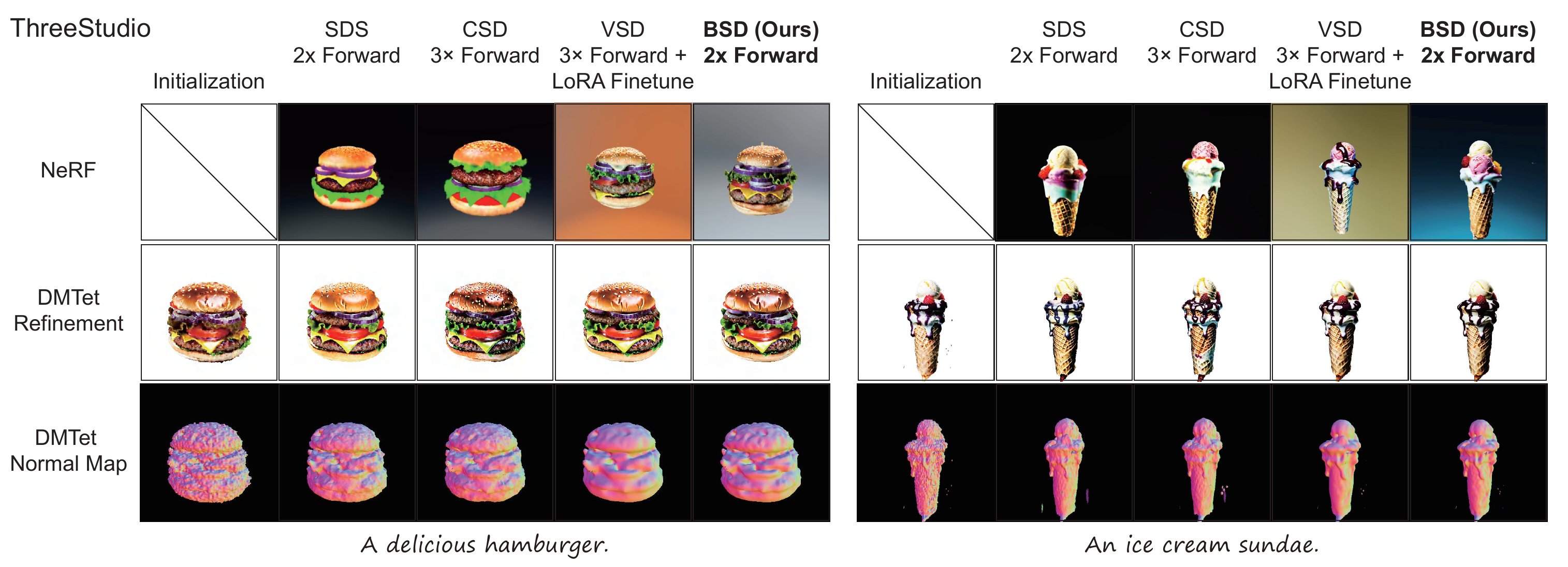}}
\vskip -0.1in
\caption{Results on ThreeStudio comparing score distillation algorithms for basic NeRF generation and DMTet refinement.}
\label{fig:BSD_threestudio_exp}
\end{center}
\vskip -0.1in
\end{figure*}
\begin{figure*}[ht]
\begin{center}
\centerline{\includegraphics[width=\textwidth]{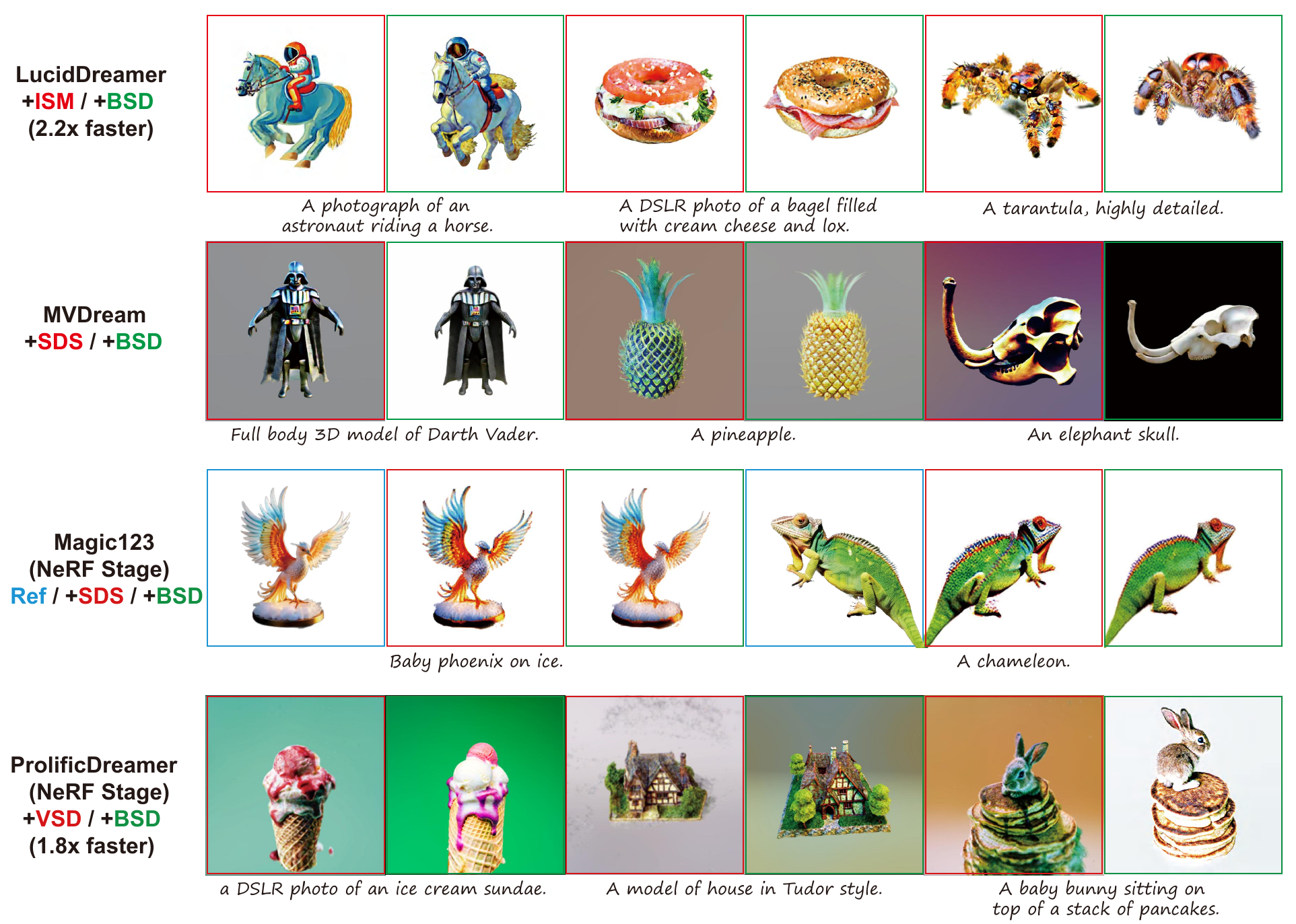}}
\vskip -0.1in
\caption{Results of replacing other score distillation methods with BSD in text-to-3D pipelines.}
\label{fig:BSD_3D_exp}
\end{center}
\vskip -0.1in
\end{figure*}
\begin{figure*}[ht]
\begin{center}
\centerline{\includegraphics[width=\textwidth]{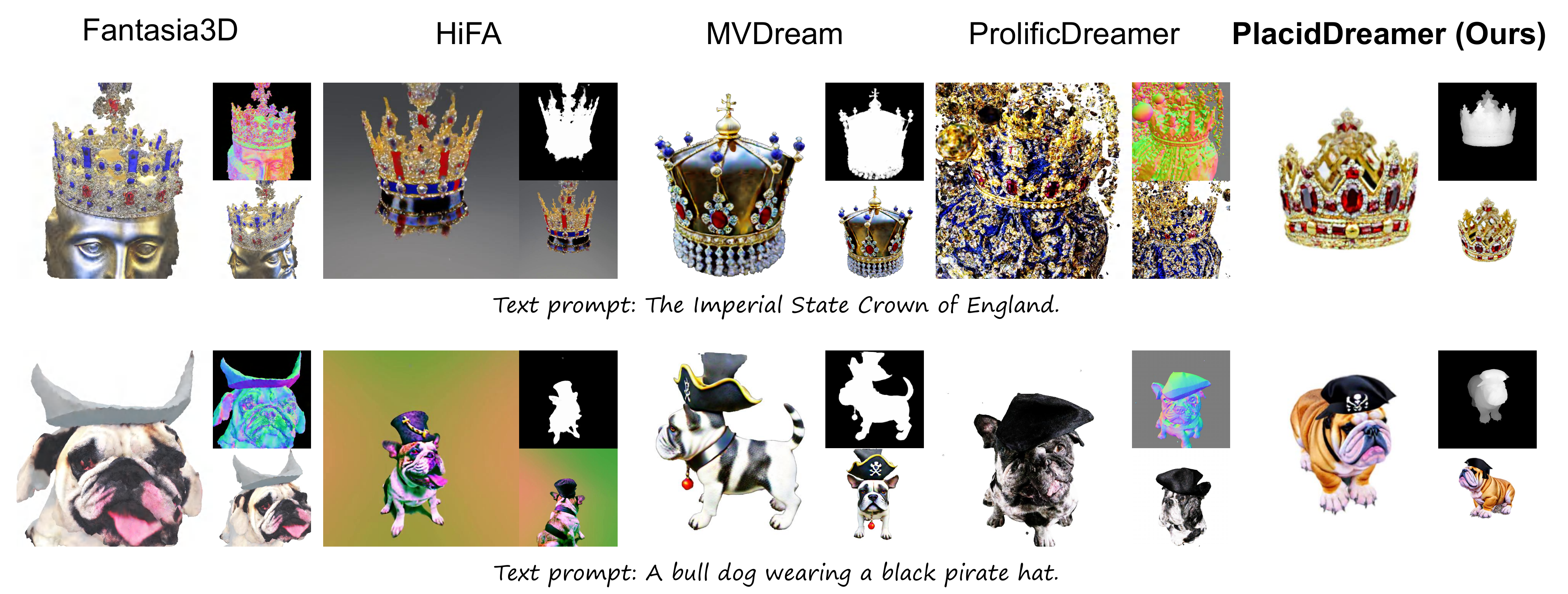}}
\vskip -0.1in
\caption{Qualitative comparison with NeRF-based baseline methods.}
\label{fig:baseline_nerf}
\end{center}
\vskip -0.1in
\end{figure*}

\section{Experiments of BSD Algorithm}
\label{sec:appendix_BSD_exp}
BSD is a versatile score distillation algorithm. We demonstrate its stability and versatility by applying it across various text-to-3D open-source frameworks. Initially, within the general framework of ThreeStudio \cite{threestudio2023}, we compare the generation capabilities of SDS \cite{poole2022dreamfusion}, CSD \cite{yu2023text}, VSD \cite{wang2023prolificdreamer}, and BSD for NeRF \cite{mildenhall2021nerf} and DMTet \cite{shen2021deep}. Subsequently, we substitute the score distillation algorithms in different frameworks with BSD, ensuring that all other parameters in the experiments remains constant, to validate its enhancements.

As illustrated in Figure~\ref{fig:BSD_threestudio_exp}, we present the experimental results on ThreeStudio. To ensure a fair comparison, we maintain all parameters that are unrelated to score distillation at constant values, while meticulously adjusting hyper-parameters of each score distillation algorithm to achieve optimal effects. In the generation of NeRFs, SDS exhibits the least detail levels and suffers from unrealistic colors. CSD, comparing with SDS, captures more pronounced details. However, the over-saturation curtails its realism. The performances of VSD and BSD are closely matched, with VSD displaying finer details and BSD displaying more accurate color distributions. BSD stands out in color accuracy, nearly matches VSD in detail richness, and equals SDS in speed, thus positioning BSD as the superior choice over previous score distillation methods. For DMTet refinement, we use a NeRF sample generated by BSD as mesh initialization. We observe that these algorithms exhibit similar quality with SDS and CSD having subtle color deviations.

As shown in Figure~\ref{fig:BSD_3D_exp}, replacing previous score distillation algorithms with the BSD algorithm consistently improves performance across various frameworks. In the LucidDreamer \cite{EnVision2023luciddreamer} pipeline, which employs Interval Score Matching (ISM), the BSD algorithm not only trains 2.2 times faster but also significantly enhances semantic alignment and texture fidelity. ISM prioritizes guidance at lower timesteps by assigning greater weights. Without resolving the conflict between classifier and smoothing guidance, this leads to over-saturation in certain scenarios (e.g., the astronaut case). Moreover, optimizations at low timesteps often add irrelevant details, enhancing visual complexity but sometimes causing semantic inconsistencies with the text prompt (e.g., the bagel case). In the MVDream \cite{shi2023mvdream} pipeline with the SDS algorithm, BSD effectively reduces the color distortion problems. The capability of BSD to reduce over-saturation is further demonstrated in experiments conducted on the Magic123 \cite{qian2023magic123} pipeline, which also show its applicability to multi-view diffusion models \cite{liu2023zero}. Furthermore, experiments on ProlificDreamer \cite{wang2023prolificdreamer} reveal that BSD not only runs much faster but also matches VSD in detail level. The experiments validate BSD as a versatile and robust choice for score distillation.

\section{Comparison with Baseline Methods}
\label{sec:appendix_qualitative}
We conduct qualitative comparisons with several NeRF-based \cite{mildenhall2021nerf} baseline methods, including Fantasia3D \cite{chen2023fantasia3d}, HiFA \cite{zhu2023hifa}, MVDream \cite{shi2023mvdream}, and ProlificDreamer \cite{wang2023prolificdreamer}. As shown in Figure~\ref{fig:baseline_nerf}, we present the results of these methods generating responses to the same prompts. Fantasia3D utilizes a unique geometry generation process for enhanced geometry generation, yet it does not match the overall quality of PlacidDreamer. MVDream is capable of generating stable, multi-view consistent 3D models. However, its training process reduces resolution, resulting in the loss of high-frequency details. ProlificDreamer produces meshes with high-fidelity textures, but it sometimes fails to converge, suffers from severe multi-face problems, and incurs a time cost significantly exceeding other methods. Our PlacidDreamer, with improvements in BSD and pipeline design, achieves balanced saturation, refined details, and more stable generation. Therefore, compared with previous NeRF-based methods, PlacidDreamer is capable of generating higher-quality 3D assets.

\section{Relationship with Previous Methods}
\label{sec:appendix_score_distillation}
\begin{figure}[ht]
\begin{center}
\centerline{\includegraphics[width=\linewidth]{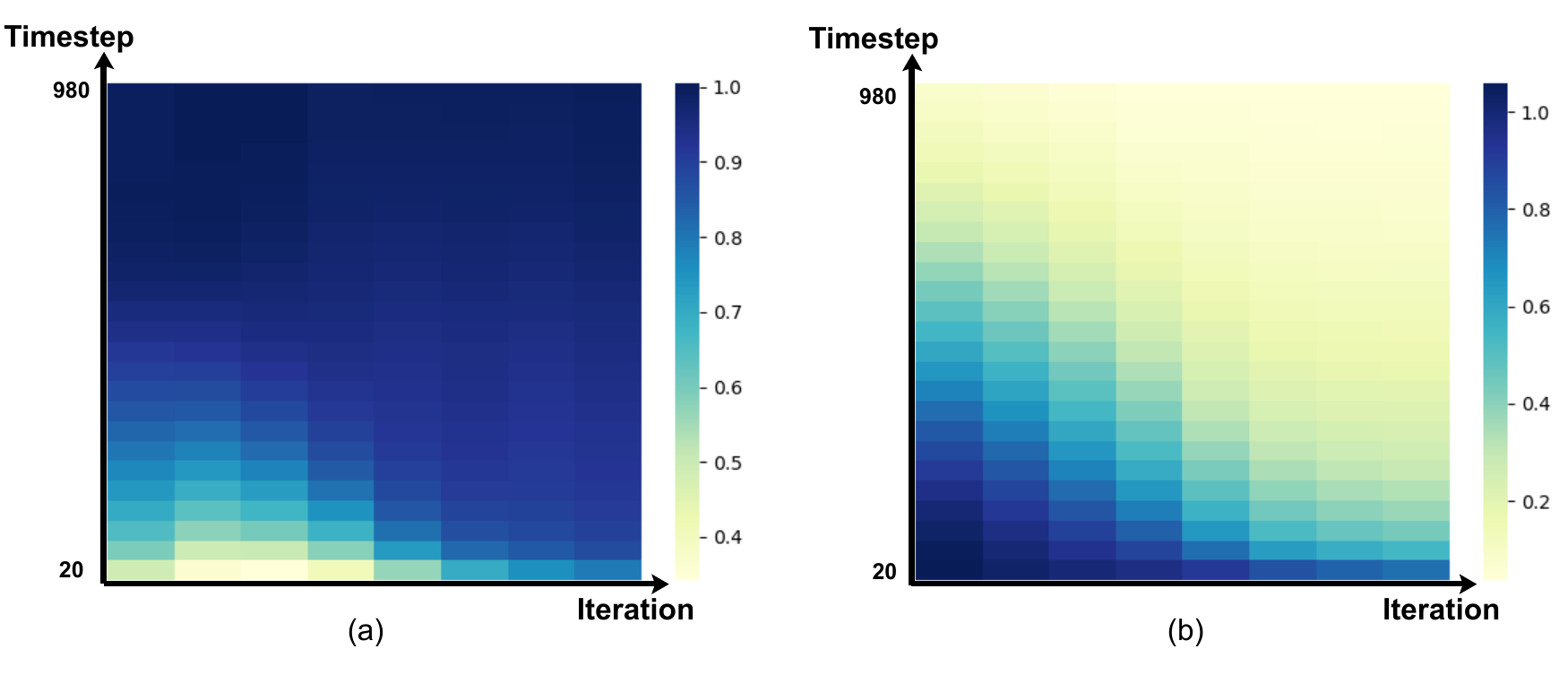}}
\vskip -0.1in
\caption{Visualization of the Euclidean norm of smoothing guidance during training: (a) $\delta_\mathrm{SG} = \epsilon(\mathbf{x}_t, t, \emptyset)$ of BSD. (b) $\delta'_\mathrm{SG}=\epsilon(\mathbf{x}_t, t, \emptyset)-\epsilon$ of SDS.}
\label{fig:variance_shift}
\end{center}
\vskip -0.1in
\end{figure}
\noindent \textbf{Comparison with SDS.}
The main differences between our approach (BSD) and SDS lie in two aspects. The first is the incorporation of the Multiple-Gradient Descent Algorithm (MGDA), as elaborated in the main text. The second is the omission of the final term $-\epsilon$. The formula we derived is similar to the one provided in the appendix of the DreamFusion \cite{poole2022dreamfusion} paper. The appendix claims that introducing $-\epsilon$ helps to reduce high variance of the gradients. Despite the common inclusion of $-\epsilon$ in most previous works that follow the SDS paradigm, we have observed that $-\epsilon$ can be omitted for three reasons.

Firstly, the introduction of $-\epsilon$ causes the magnitude of $\delta'_\mathrm{SG}=\epsilon(\mathbf{x}_t, t, \emptyset) - \epsilon$ exhibit greater variance across different timesteps compared with $\delta_\mathrm{SG}=\epsilon(\mathbf{x}_t, t, \emptyset)$. As we can clearly see in Figure~\ref{fig:variance_shift}, the color blocks within each column of $\delta_\mathrm{SG}$ are almost identical, indicating that their magnitudes are similar. In contrast, the colors in each column of $\delta'_\mathrm{SG}$ change significantly with the timestep, which indicates that their magnitudes vary widely. This results in a fixed CFG value being more difficult to control balance. Additionally, we find that $\delta'_\mathrm{SG}$ has a higher likelihood of producing the obtuse angles between two optimization directions at low timesteps. These factors result in a more severe over-saturation problem. 

Secondly, the introduction of $-\epsilon$ causes the magnitude of $\delta'_\mathrm{SG}$ to decrease during optimization, disrupting the balance. As the rendered 2D images progressively mirror the distribution of real 2D images, the predicted noise $\epsilon(\mathbf{x}_t, t, \emptyset)$ and the added random noise $\epsilon$ become numerically correlated. This correlation lowers the magnitude of $\delta'_\mathrm{SG}$, visibly lightening the color blocks in each row in Figure~\ref{fig:variance_shift} (b). Consequently, in this scenario, classifier guidance is likely to dominate when the CFG parameter is fixed, leading to a more pronounced over-saturation issue. When MGDA is applied, its mathematical property—increasing the proportion of components as their magnitudes decrease—causes $\delta'_\mathrm{SG}$ to likely dominate in the MGDA algorithm, exacerbating the over-smoothing problem. Thus, the magnitude reductions caused by $-\epsilon$ are detrimental to achieving balance.

Thirdly, despite $-\epsilon$ having a mathematical expectation of zero, practical challenges arise. Text-to-3D algorithms typically run only a few thousand to tens of thousands of iterations. With 1000 different timesteps involved, the few dozen random samples per timestep are insufficient to mitigate their mutual impacts effectively.

\begin{figure}[ht]
\begin{center}
\centerline{\includegraphics[width=\linewidth]{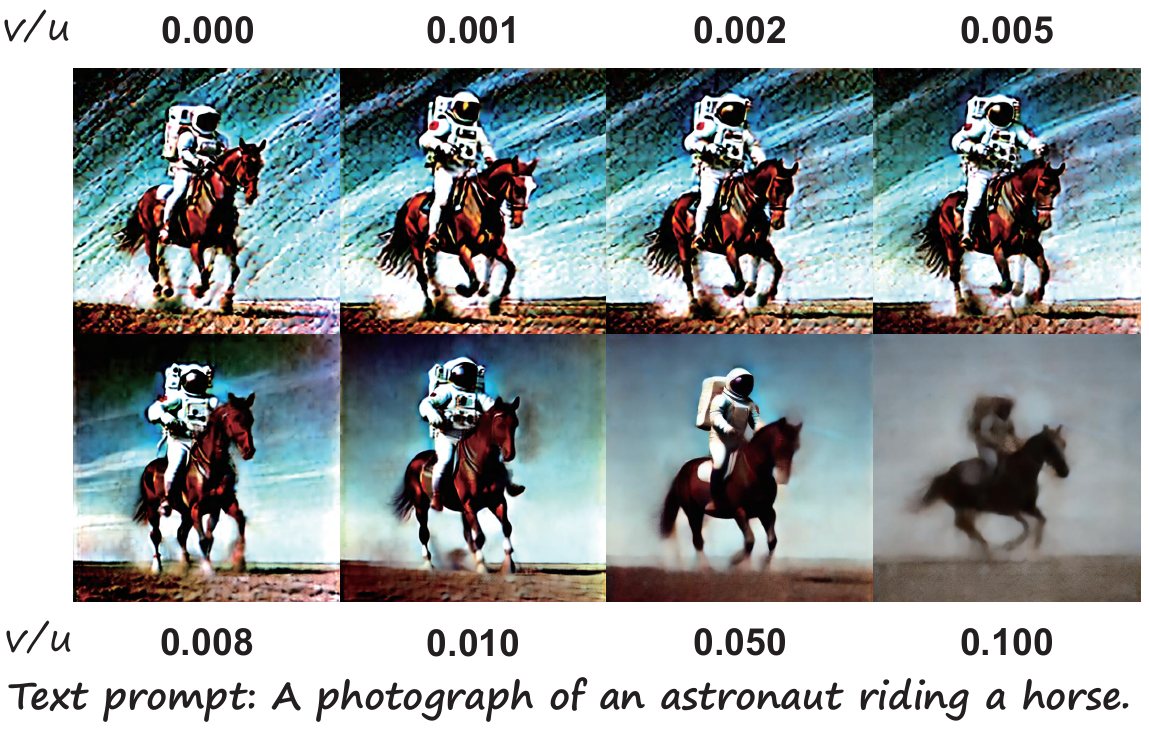}}
\vskip -0.1in
\caption{The influence of the ratio between smoothing guidance and classifier guidance, represented as $v/u$ ($-\nabla_{\mathbf{x}_t} \log p_t(\mathbf{x}_t|y) = u \cdot \delta_\mathrm{CG} + v \cdot \delta_\mathrm{SG}$). When the ratio $v/u$ is set to zero, it corresponds to CSD.}
\label{fig:decomposition}
\end{center}
\vskip -0.1in
\end{figure}

\noindent \textbf{Comparison with CSD.}
The decomposition method of CSD is similar to ours, as both include a classifier guidance term $\delta_\mathrm{CG}$. However, their $\delta^\mathrm{gen} = \epsilon(\mathbf{x}_t, t, y) - \epsilon$, whereas our $\delta_\mathrm{SG} = \epsilon(\mathbf{x}_t, t, \emptyset)$. In most cases, experimental results using $\epsilon(\mathbf{x}_t, t, y)$ and $\epsilon(\mathbf{x}_t, t, \emptyset)$ do not differ significantly. It should be noted that at the initial stages of 3D generation, when initialized to a sphere, the image is smooth, thus $\epsilon(\mathbf{x}_t, t, \emptyset)$ is correlated with $\epsilon$, while $\epsilon(\mathbf{x}_t, t, y)$ is not correlated with $\epsilon$. This means that the $\epsilon(\mathbf{x}_t, t, \emptyset)$ is more closely related to smoothness. Therefore, it is more accurate to refer to our decomposition as the 'smoothing guidance'.

A more significant difference arises in understanding the effects of the decomposed terms. CSD discovers that $\delta^\mathrm{gen}$ alone does not function effectively and occupies a very low proportion in terms of magnitude, leading to the conclusion that $\delta^\mathrm{gen}$ can be discarded. However, according to our modeling, using only $\delta_\mathrm{CG}$ causes the generated results to overfit to a mode learned by a classifier, thereby distilling discriminative power but not utilizing generative capabilities. This leads to noticeable artifacts in 2D experiments and over-saturation issues in 3D experiments produced by CSD. We display the results of the score distillation decomposition experiments in higher detail in Figure~\ref{fig:decomposition}. Furthermore, we believe that precisely because the proportion of $\delta_\mathrm{SG}$ is very small, the MGDA algorithm is necessary to ensure its effects are not overwhelmed, hence we propose the BSD algorithm.

\begin{figure}[h]
\begin{center}
\centerline{\includegraphics[width=\linewidth]{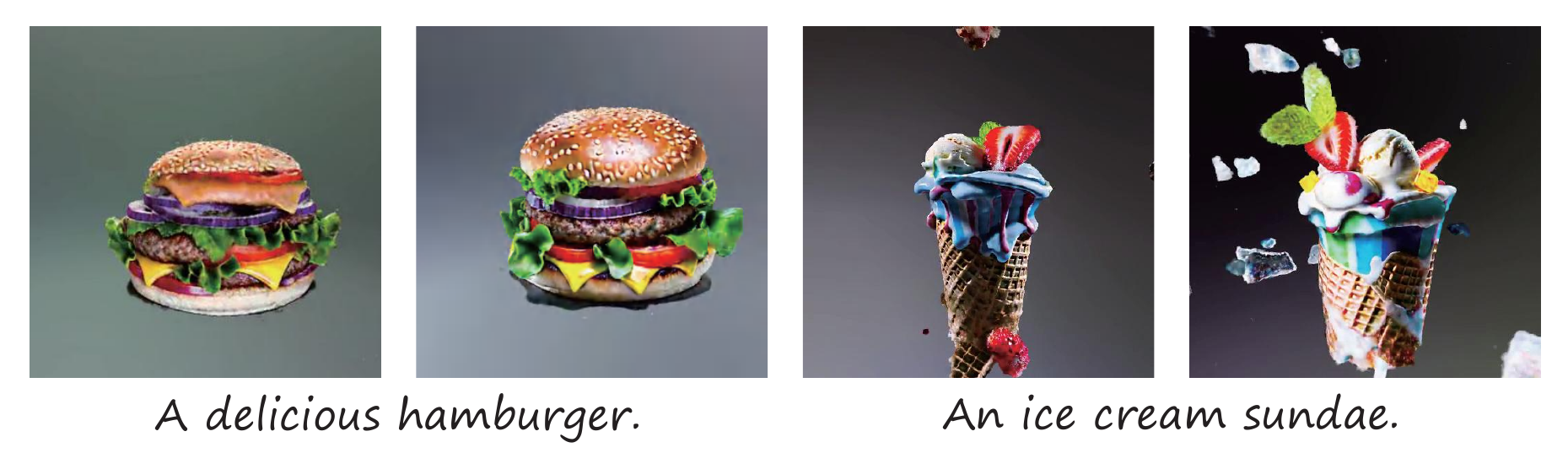}}
\vskip -0.1in
\caption{The results of applying MGDA to VSD.}
\label{fig:MGDA_VSD}
\end{center}
\vskip -0.1in
\end{figure}

\noindent \textbf{Comparison with VSD.}
The derivation approach for VSD involves sampling from a 3D distribution as part of variational inference and utilizing a particle-based ODE (Ordinary Differential Equation) solvor. The final expression derived is:
$$\nabla_\theta{\mathcal{L}_\mathrm{VSD}(\theta)}=\mathbb{E}_{t,\epsilon,c}[\omega(t)(\epsilon_\mathrm{pretrain}(\mathbf{x}_t, t, y) - \epsilon_\phi(\mathbf{x}_t, t, y))\frac{\partial g(\theta, \pi)}{\partial \theta}],$$
where $\theta$ represents the parameters of the 3D model, $\epsilon_\mathrm{pretrain}$ is the noise predicted by the pre-trained diffusion model, $\epsilon_\phi$ is the noise predicted by the diffusion model fine-tuned on the optimizing 3D assets, and $g(\theta, \pi)$ is the differentiable renderer for the 3D assets from perspective $\pi$.

Unlike SDS, VSD cannot be decomposed into a combination of classifier guidance and smoothing guidance clearly, because $\epsilon_\phi(\mathbf{x}_t, t, y)$ inherently carries probabilistic meanings, complicating its integration with any guidance term. However, we still test the effects of incorporating $-\epsilon_\phi(\mathbf{x}_t, t, y)$ into the smoothing guidance and applying MGDA. The results, as illustrated in Figure~\ref{fig:MGDA_VSD}, demonstrate improvements in detail level and content richness. Since VSD modeling is not the central focus of our paper, we leave the potential exploration of integrating VSD with MGDA to future work.

\section{Implementation Details}
\label{sec:appendix_implementation_details}

In the Latent-Plane module, we utilize two Multi-Head Self-Attention layers, each with eight heads and a feature dimension of 32, to extract sigma features. An additional two layers are used to enhance the multi-view features. For embeddings not derived from neural networks, we employ sinusoidal encoding. To minimize computational demands, all MLP networks are comprised of a single linear layer. During LoRA finetuning, only the UNet LoRA layers are fine-tuned at a learning rate of $1 \times 10^{-4}$ across 400 iterations. In our BSD implementation, we set $\lambda=25$ and $\omega(t) \propto \alpha_t^2$ to achieve an optimal balance of rich detail and accurate color reproduction.


\end{document}